\documentclass{article}
\usepackage{multirow}
\usepackage{multicol}
\usepackage{tabu}
\usepackage{graphicx}
\usepackage{makecell}
\usepackage{amsthm}
\usepackage{mathtools}
\usepackage{microtype}
\usepackage[export]{adjustbox}
\usepackage[T1]{fontenc}
\usepackage[scaled=0.825]{beramono}
\usepackage{hhline}
\usepackage[ruled,vlined,linesnumbered,algo2e]{algorithm2e}
\usepackage{xspace}
\usepackage{lipsum}
\usepackage{comment}
\usepackage[super]{nth}
\usepackage{array}
\usepackage{balance}

\usepackage{amsmath,amsfonts}
\usepackage{algorithmic}
\usepackage{textcomp}
\usepackage[table]{xcolor}
\usepackage{listings, color}
\usepackage{tikz}
\usepackage{comment}
\usepackage[bb=boondox]{mathalfa}
\usepackage[figuresright]{rotating}

\newcommand{\para}[1]{\noindent \textbf{#1.}}

\newenvironment{itemizeReduced}{
\begin{list}{\labelitemi}{\leftmargin=1em}
\setlength{\itemsep}{1pt}
\setlength{\parskip}{0pt}
\setlength{\parsep}{0pt}}{\end{list}
}
\definecolor{mauvelous}{rgb}{0.94, 0.6, 0.67}
\definecolor{mauvetaupe}{rgb}{0.57, 0.37, 0.43}
\lstdefinestyle{myStyle}{
  belowcaptionskip=1\baselineskip,
  frame=tb,
  language=python,
  aboveskip=0mm,
  belowskip=0mm,
  showstringspaces=false,
  columns=flexible,
  basicstyle={\fontsize{8pt}{8pt}\fontfamily{fvm}\selectfont},
  numbers=left,
  xleftmargin=3em,
  numberstyle={\color{gray}\texttt},
  keywordstyle=\color{black}\textbf,
  commentstyle=\color{mauvelous},
  frame=none,
  breaklines=true,
  breakatwhitespace=true,
  tabsize=2,
  morekeywords={where},
  keywords={[2]{}},
}

\renewenvironment{itemizeReduced}{
\begin{list}{\labelitemi}{\leftmargin=1em}
\setlength{\itemsep}{1pt}
\setlength{\parskip}{0pt}
\setlength{\parsep}{0pt}}{\end{list}
}

\newcommand*\circled[1]{\tikz[baseline=(char.base)]{
            \node[shape=circle,fill,inner sep=1pt] (char) {\textcolor{white}{#1}};}}

\usepackage{microtype}
\usepackage{graphicx}
\usepackage{subfigure}
\usepackage{booktabs} 

\usepackage{hyperref}

\usepackage[accepted]{icml2024}

\usepackage{amsmath}
\usepackage{amssymb}
\usepackage{mathtools}
\usepackage{amsthm}

\usepackage[capitalize,noabbrev]{cleveref}

\theoremstyle{plain}

\theoremstyle{definition}

\theoremstyle{remark}

\icmltitlerunning{Any-Precision LLM: Low-Cost Deployment of Multiple, Different-Sized LLMs}

\begin{document}

\twocolumn[
\icmltitle{Any-Precision LLM: Low-Cost Deployment of Multiple, Different-Sized LLMs}



\icmlsetsymbol{equal}{*}

\begin{icmlauthorlist}
\icmlauthor{Yeonhong Park}{yyy}
\icmlauthor{Jake Hyun}{yyy}
\icmlauthor{SangLyul Cho}{yyy}
\icmlauthor{Bonggeun Sim}{yyy}
\icmlauthor{Jae W. Lee}{yyy}
\end{icmlauthorlist}

\icmlaffiliation{yyy}{Seoul National University}

\icmlcorrespondingauthor{Jae W. Lee}{jaewlee@snu.ac.kr}


\vskip 0.3in
]



\printAffiliationsAndNotice{}  

\begin{abstract}
Recently, considerable efforts have been directed towards compressing Large Language Models (LLMs), which showcase groundbreaking capabilities across diverse applications but entail significant deployment costs due to their large sizes. Meanwhile, much less attention has been given to mitigating the costs associated with deploying multiple LLMs of varying sizes despite its practical significance. Thus, this paper introduces \emph{any-precision LLM}, extending the concept of any-precision DNN to LLMs. Addressing challenges in any-precision LLM, we propose a lightweight method for any-precision quantization of LLMs, leveraging a post-training quantization framework, and develop a specialized software engine for its efficient serving. As a result, our solution significantly reduces the high costs of deploying multiple, different-sized LLMs by overlaying LLMs quantized to varying bit-widths, such as 3, 4, ..., $n$ bits, into a memory footprint comparable to a single $n$-bit LLM. All the supported LLMs with varying bit-widths demonstrate state-of-the-art model quality and inference throughput, proving itself to be a compelling option for deployment of multiple, different-sized LLMs. The code is available at \url{https://github.com/SNU-ARC/any-precision-llm}.
\end{abstract}

\section{Introduction}
\label{sec:intro}
With the revolutionary success of Large Language Models (LLMs) across various applications, there have been many recent efforts to reduce the costs of their deployment. Specifically, there has been a considerable focus on compressing LLMs using techniques like pruning~\cite{prune-1,prune-2,prune-3,prune-4} or quantization~\cite{awq,gptq,sqllm}, as the parameter size is the primary obstacle for their efficient deployment. 

Meanwhile, there has been limited discussion on mitigating the costs associated with deploying multiple LLMs of varying sizes, despite its practical significance. Real-world scenarios often demand the dynamic adaptation of multiple LLMs, each with distinct model quality/inference latency trade-offs. This approach enables the effective handling of queries with varied latency constraints, enhancing the user experience. Moreover, it supports a popular generation acceleration technique: speculative decoding~\cite{speculative-1, speculative-2, speculative-3}. Despite these benefits, deploying multiple LLMs of varying sizes presents challenges. First, it exacerbates the already high memory costs of LLM deployment, and second, it necessitates training of multiple model versions when models of desired sizes are not readily available as open-source.

Any-precision LLM, an extension of the concept of any-precision DNN~\cite{any-precision} to LLM, is a promising solution for the low-cost deployment of multiple, different-sized LLMs. Any-precision DNN refers to an $n$-bit quantized model capable of generating lower bit quantized models (($n$-1)-bit, ($n$-2)-bit, ...) simply by taking its most significant bits (MSBs). Applying this concept to LLM enables the utilization of multiple LLMs with varying sizes by storing only a single large LLM ($n$-bit model) in memory, while avoiding the additional overhead of training multiple LLMs.

Meanwhile, two challenges need to be resolved for effective implementation of any-precision LLM. First, a practical method for any-precision quantization of LLM is needed. The existing any-precision quantization method on DNN requires training the model from scratch, limiting its applicability to LLMs. Second, a new GPU kernel for quantized matrix-vector multiplication is required, which will translate the use of reduced bit-widths in any-precision LLMs into shorter inference times. Existing kernels for quantized matrix-vector multiplication are unable to load just a portion of each quantized weight value's bit-vector. Consequently, with existing kernels, opting for a model with a lower bit-width does not reduce memory bandwidth usage.

Thus, we in this paper make a strong case for any-precision LLM by addressing the two aforementioned issues. First, we build a lightweight method for any-precision quantization of LLM. Utilizing a post-training quantization (PTQ) framework, it first generates a low-bit model and then incrementally upscales it to higher bit-widths, conserving the any-precision property. Second, we develop a new software engine specialized for any-precision support, that effectively saves memory bandwidth for serving any-precision LLMs by changing the memory layout of weights. Our extensive experimental studies demonstrate that our solution is a powerful approach for the deployment of multiple, different-sized LLMs, achieving the following results:

\begin{itemizeReduced}
    \item Our solution efficiently packs LLMs quantized to varying bit-widths, such as 3, 4, ... up to $n$ bits, into a memory footprint comparable to a single $n$-bit LLM.
    \item Our solution yields a set of quantized LLMs of varying bit-widths that, while offering any-precision support, match the quality of the state-of-the-art quantization techniques at each bit-width.
    \item Our solution, despite having to adopt a bit-interleaved (bitplane) memory layout for the support of any-precision, showcases high inference throughput, matching or even outperforming that of state-of-the-art quantized matrix-vector multiplication engines that do not support any-precision~\cite{sqllm}.
\end{itemizeReduced}



\section{Background}
\label{sec:back}
\subsection{GPU Basics}
\para{GPU Architecture Fundamentals} 
A GPU, the de facto standard platform for executing LLMs, comprises a large number of processing elements called Streaming Multiprocessors (SMs). 
GPUs often include multi-level on-chip SRAM caches. A part of the L1 cache can be configured as shared memory, providing a memory space that can be directly controlled by programmers.


\para{Execution Model}
GPUs use a large number of threads to execute operations, which are known as kernels. Threads are structured into thread blocks whose execution is scheduled on SMs. All threads within a thread block share the same shared memory space. Within each thread block, threads are further organized into a set of warps, with each warp consisting of 32 consecutive threads. All threads within a warp execute the same instruction at the same time. 

\subsection{LLM Quantization}
\label{sec:algo:prior-works}
This section discusses recent advancements in LLM quantization with a specific emphasis on weight-only, post-training quantization (PTQ). In LLMs, there is a pronounced shift towards quantizing only the weights, as the dominant bottleneck in inference throughput is the memory constraint imposed by the size of weight parameters, rather than computational requirements. 
Moreover, Post-Training Quantization (PTQ) has become a favored method for quantizing LLMs due to its practicality. Although Quantization-Aware Training (QAT) typically yields superior performance, its high training expense frequently renders it impractical~\cite{llmqat, qlora, peqa}.


GPTQ~\cite{gptq}, a pioneering work on weight-only PTQ for LLM, formulates quantization as a layer-wise weight reconstruction problem. GPTQ methodically quantizes each channel in iterations, simultaneously adjusting the remaining not-yet-quantized weights to correct for quantization-induced errors. AWQ~\cite{awq} performs per-channel scaling to safeguard a small fraction of salient weights as a preprocessing step. Similarly, QuIP~\cite{quip} preprocesses weights to be more amenable to quantization, yielding impressive results even at 2-bit precision. However, its practical utility is in question as it adds a substantial runtime overhead.

While the aforementioned methods employ uniform quantization, non-uniform quantization may be a more effective alternative as it better captures the weight distributions~\cite{quant-survey}. SqueezeLLM~\cite{sqllm} proposes a clustering-based LLM quantization that considers the sensitivity of each weight. Somewhat orthogonal to the aforementioned studies that primarily focus on rounding schemes, there are also proposals to use mixed precision as a way of allocating more bits to sensitive weights~\cite{sqllm, spqr, owq}.

\section{Motivation}
\label{sec:motiv}
\subsection{Need for Deploying Multiple, Different-sized LLMs}
Deploying a set of different-sized LLMs provides significant practical advantages. It enhances user experience by effectively handling queries with varying latency requirements. Depending on the specific needs of users and applications, some queries require quick responses, while others can tolerate slower response times. Latency requirements become even more diverse when serving multiple different tasks concurrently, a common use case of LLMs. For instance, queries for interactive tasks like chatbots are mostly latency-sensitive, while tasks like document analysis, often handled in the background, allow for more relaxed response times. In fact, the approach of adaptively selecting different DNN models with various accuracy-latency trade-offs has been widely studied as a way to effectively meet user requirements due to its practical significance~\cite{adapt-1, adapt-2, adapt-3}.

Another scenario that necessitates multiple LLMs of varying sizes is speculative decoding. This popular technique boosts the throughput of a large model by additionally utilizing one or more smaller draft models~\cite{speculative-1, speculative-2, speculative-3, speculative-4}. 

\subsection{Challenges of Deploying Multiple, Different-sized LLMs}
In deploying multiple LLMs, we face two practical challenges: memory overhead and training costs.

\para{Challenge 1: Memory Overhead}
First, a large memory capacity overhead is introduced. As LLMs are typically substantial in size, maintaining even a few additional smaller models incurs significant costs. For example, deploying three tiers of LLMs with varying sizes --- a large base model, a half-sized model, and a quater-sized model --- nearly doubles on the total memory requirement.

\para{Challenge 2: Training Costs}
Acquiring multiple different-sized LLMs is challenging in itself. While some open-source LLMs such as OPT~\cite{opt} offer a comprehensive range of models with varying parameter counts, this is usually not the case. Most open-source LLMs offer only one to three variants, limiting their versatility across different use scenarios. If a model in the desired size is unavailable, users must create it by themselves. However, training an LLM is very costly due to its high computational needs and large corpus requirement. One approach is to distill the available large model into smaller ones instead of training smaller models from scratch~\cite{distill-1, distill-2, distill-3, distill-4, distill-5}. While this approach reduces computation costs, it still entails non-trivial engineering challenges, including the need to assemble a proper set of training data and finding favorable hyperparameters for training.

These two challenges become particularly prominent when individuals run LLMs on personal platforms like desktops and mobile devices. In these scenarios --- unlike with datacenter-scale inference --- compute, memory, and even engineering resources are severely limited. Thus, this paper focuses on addressing the challenges associated with deploying multiple, different-sized LLMs for on-device inference.

\begin{figure}[t]
    \centering
    \includegraphics[width=\linewidth]{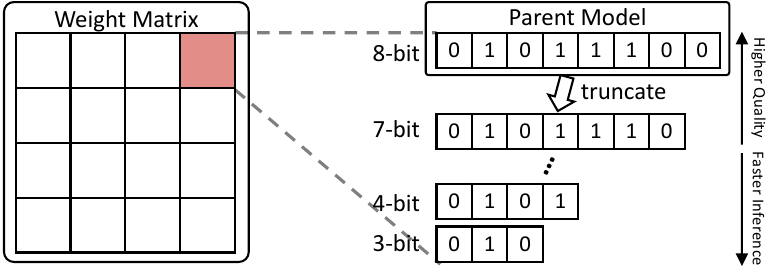}
    \caption{Concept of any-precision quantization.}
    \label{fig:any-precision}
\end{figure}

\begin{table}[t]
\caption{Memory savings of any-precision LLM when deployed on various sets of bit-widths for Llama-2-7B.}
\label{table:saving}
\resizebox{\linewidth}{!}{%
\begin{tabular}{cccc}
\begin{tabular}[c]{@{}c@{}}Supported \\ Bit-widths\end{tabular} & \begin{tabular}[c]{@{}c@{}}Any-Precision \\ LLM\end{tabular} & \begin{tabular}[c]{@{}c@{}}Separate \\ Deployment\end{tabular} & \begin{tabular}[c]{@{}c@{}}Memory \\ Savings\end{tabular} \\ \hline
\{3,6\}                                                           & 5.6 GB                                                       & 8.3 GB                                                         & 1.49$\times$                                                     \\
\{4,8\}                                                           & 7.7 GB                                                       & 10.8 GB                                                         & 1.40$\times$                                                     \\
\{3,4,6\}                                                           & 5.6 GB                                                       & 12.1 GB                                                         & 2.15$\times$                                                     \\
\{3,4,8\}                                                         & 7.7 GB                                                       & 13.7 GB                                                         & 1.76$\times$                                                    \\
\{3,4,6,8\}                                                         & 7.9 GB                                                       & 19.1 GB                                                         & 2.41$\times$                                                      \\
\{3,4,5,6,7,8\}                                                   & 8.4 GB                                                       & 29.9 GB                                                         & 3.56$\times$                                                    
\end{tabular}%
}
\end{table}

\subsection{Our Solution: Any-Precision LLM}
\para{Concept} Any-precision quantization, initially introduced in prior work~\cite{any-precision}, is a promising solution to mitigate the costs of deploying multiple different-sized LLMs. Figure~\ref{fig:any-precision} visualizes the concept of any-precision quantization. The core is to derive smaller models (7-bit, 6-bit quantized model, ...) from a large model (8-bit quantized model), referred to as the parent model, by taking only the upper bits of its parameters. Of course, a special method tailored for any-precision quantization is required to ensure that taking just the prefixes of the parent model parameters does not result in significant quality drops. The any-precision approach is highly memory-efficient as it allows the utilization of varying bit-width models by only storing in memory 1) the quantized weights of the parent model and 2) a set of quantization parameters (e.g. centroid values) associated with each supported bit-width, relatively small in size. Also, there is no need to train multiple models.

Table~\ref{table:saving} gives a quantitative analysis on the memory savings of any-precision LLM, the application of any-precision quantization to LLMs. Without any-precision quantization, adaptively using models of different bit-widths requires the deployment of separate models, each taking up its own memory space. We refer to this strategy as separate deployment. We compare the required memory space of separate deployment against any-precision LLM, for various scenarios requiring different sets of bit-widths with the Llama-2-7B model. Any-precision LLM significantly reduces memory costs across a range of scenarios, achieving a maximum saving of 3.56$\times$ when supporting all bit-widths from 3 to 8.

\para{Challenges of Any-Precision LLM} While the concept itself is appealing, there are critical issues in directly applying the method of the original work~\cite{any-precision}, which mainly targets CNN models, to LLMs. First, it is a quantization-aware training (QAT) scheme, requiring models to be trained from scratch. During training, the forward pass quantizes parameters to varying bit-widths so that the resulting model is robust to any-precision quantization. However, for LLMs, training is not affordable to most users. Second, this work has no regard for memory bandwidth saving. The entire $n$-bit parameters of the parent model are loaded into memory, only then being further quantized into lower bit-width weights by bit-shifting as needed. This strategy makes sense for CNN models, which are usually compute-bound. On the other hand, on-device LLM inference is highly memory-bound due to its low arithmetic intensity. Consequently, the memory load of weight parameters is the single primary performance bottleneck~\cite{full-stack}. Hence, when the original method is applied to LLMs as-is, operating at low bit-widths may provide little inference latency improvements over that of the parent model. A new solution is required for LLMs, one that incorporates both a low-cost any-precision quantization method and a specialized software engine wherein reduced precision inference directly translates to actual speed-ups.



%
\begin{figure}[t]
    \centering
    \includegraphics[width=\linewidth]{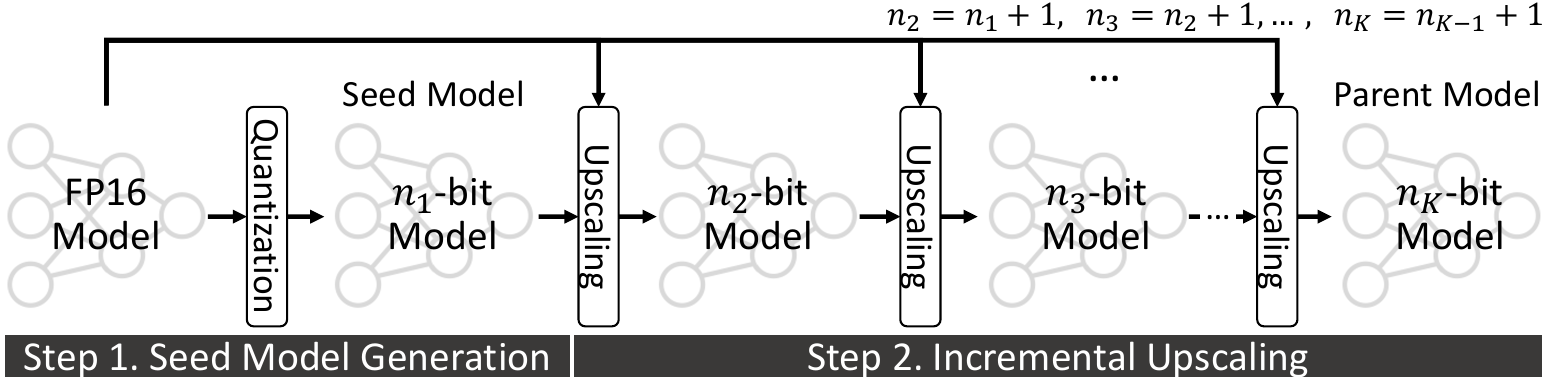}
    \caption{Any-precision quantization via incremental upscaling.}
    \label{fig:flow}
\end{figure}

\section{Any-Precision Quantization for LLM}
\label{sec:algo}

\subsection{Incremental Upscaling}

We propose a novel approach to any-precision quantization of LLMs, termed \emph{incremental upscaling}. Figure~\ref{fig:flow} illustrates the two-stage flow of the any-precision quantization, utilizing incremental upscaling. Assuming a list of candidate bit-widths ($\{{n_{k}}\}_{k=1}^{K}$), the initial step quantizes the model to the minimum supported bit-width ($n_{1}$), which we refer to as \emph{seed model}. Subsequently, we incrementally upscale the seed model one bit at a time, until we obtain the final $n_{K}$-bit parent model. For every incremental upscale from an $n_{i}$-bit model to an $n_{i+1}$-bit model, all parameters of the $n_{i}$-bit model are inherited to the $n_{i+1}$-bit model, and a single additional bit is appended to the end of each parameter. 



\begin{figure}[t]
    \centering
    \includegraphics[width=\linewidth]{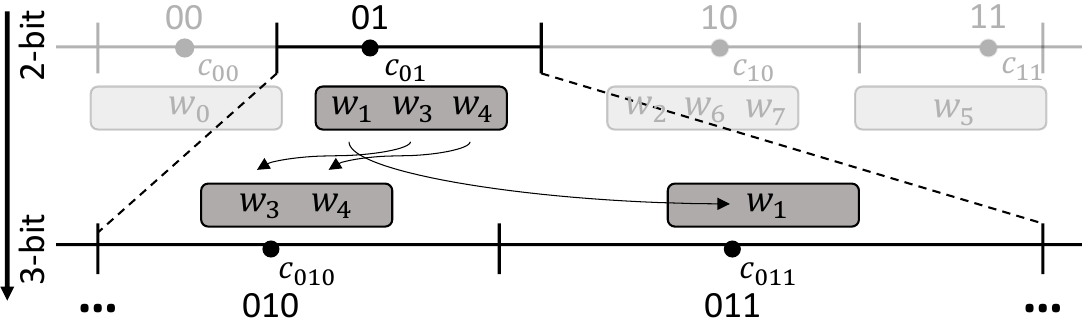}
    \vspace{-0.5cm}
    \caption{Incremental upscaling from 2 to 3-bit on non-uniform quantization methods.}
    \label{fig:upscale}
    \vspace{-0.5cm}
\end{figure}

\begin{table}[t]
\caption{Perplexity increase observed in uniform quantization methods (AWQ, GPTQ) with the application of incremental upscaling (IU). The evaluation is performed using Llama-2-7B on Wikitext2.}
\label{table:uniform}
\resizebox{\linewidth}{!}{%
\begin{tabular}{ccccccc}
        & 3-bit & 4-bit & 5-bit & 6-bit & 7-bit & 8-bit \\ \hline
AWQ     & 6.24  & 5.59  & 5.50  & 5.48  & 5.47  & 5.47  \\
AWQ+IU  &       & INF   & 22.50 & 8.16  & 6.07  & 5.67  \\ \hline
GPTQ    & 6.73  & 5.68  & 5.52  & 5.48  & 5.47  & 5.47  \\
GPTQ+IU &       & 12.7  & 6.00  & 5.79  & 5.74  & 5.73  \\ \hline
\end{tabular}%
}
\end{table}

\subsection{Non-uniform Quantization-based Incremental Upscaling}
For our incremental upscaling-based approach of any-precision quantization, a particular quantization method must be adopted as a backbone for both the seed model generation and the subsequent upscaling process. While any PTQ method can be employed for this purpose, we rule out the use of QAT method to ensure that our solution does not involve an expensive training process. Among various PTQ methods, an optimal choice would be one that 1) demonstrates state-of-the-art quantization results for the generation of a quality seed model and 2) seamlessly extends to the incremental upscaling process. To this end, we utilize the state-of-the-art clustering-based non-uniform quantization scheme, SqueezeLLM~\cite{sqllm}, as the backbone method. It delivers state-of-the-art results and is readily compatible with incremental upscaling.

Incremental upscaling with the clustering-based quantization method is straightforward. In SqueezeLLM, quantization is a process of clustering the parameters, where values within each cluster are rounded to its centroid. In this light, incremental upscaling can be achieved by further dividing each cluster into two sub-clusters. Specifically, we perform a weighted K-means clustering on the values of each cluster to generate the two sub-clusters. A sensitivity metric based on an approximated second-order derivative is used during clustering, like in the original method.

Figure~\ref{fig:upscale} visualizes the upscaling process on the clustering-based non-uniform quantization method, assuming the case of producing a 3-bit model from a 2-bit model. In this instance, three weight parameters ($w_{1}, w_{3}, w_{4}$) initially assigned to cluster \texttt{01} --- previously rounded to its centroid $c_{01}$ --- are now divided into two new sub-clusters, namely \texttt{010} and \texttt{011}, each with its respective centroids $c_{010}$ and $c_{011}$. The same process is applied to the remaining three clusters (\texttt{00}, \texttt{10}, \texttt{11}), resulting in a final total of eight sub-clusters.

In contrast to clustering-based methods, state-of-the-art \emph{uniform} quantization methods often involve complicated mechanisms such as weight reconstruction~\cite{gptq} and per-channel scaling~\cite{awq} that are not readily compatible with incremental upscaling. For example, Table~\ref{table:uniform} presents the results of applying incremental upscaling to two state-of-the-art uniform methods (AWQ, GPTQ), using a 3-bit model as the seed model. The upscaled models consistently underperform the independently quantized models, showing unacceptable degradation at low bit-widths such as 4. We provide an in-depth discussion on the challenges of applying incremental upscaling on existing uniform quantization methods, along with additional experimental results, in Appendix~\ref{appendix:uniform}.

\section{Specialized Software Engine} 
\label{sec:system}

\subsection{Need for New Software Engine}
\label{sec:system:motivation}
\begin{figure}[t]
    \centering
    \includegraphics[width=\linewidth]{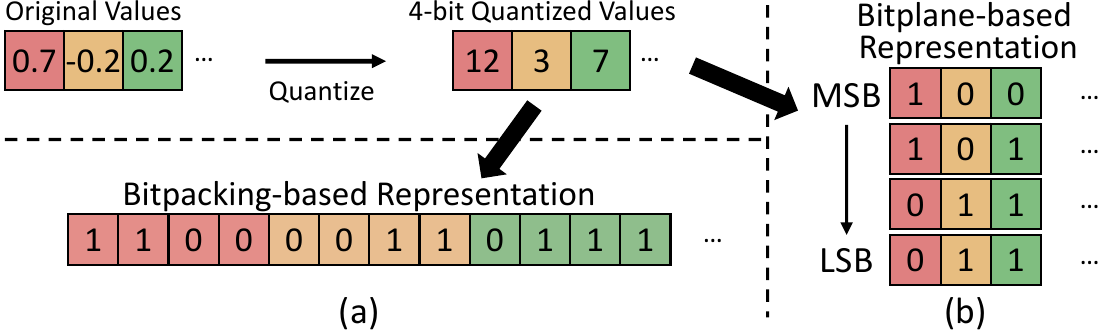}
    \caption{Comparison of (a) bitpacking-based and (b) bitplane-based representations of quantized weights.}
    \label{fig:representation}
\end{figure}

There are multiple GPU kernels designed for the efficient execution of weight-only-quantized LLMs~\cite{tensorrt-llm, awq, gptq, sqllm, exllamav2, lutgemm}. While demonstrating promising performance, most of these existing implementations cannot support any-precision quantization because of their way of representing quantized weights. They use a \emph{bitpacking}-based representation, illustrated in Figure~\ref{fig:representation}-(a). Bitpacking-based representations store quantized weights sequentially in a single 1-D array. With this representation, the entire weight array has to be loaded even when running a model in a reduced bit-width. For example in Figure~\ref{fig:representation}-(a), even when executing a 2-bit or 3-bit model, the full 4-bit values have to be read from memory. This is because of the coarse-grained memory access granularity of GPUs, typically 128 bytes. This results in virtually no performance improvements. 

On the contrary, \emph{bitplane}-based representation is a suitable choice for any-precision LLM support. Bitplane-based representation decomposes quantized weights into $n$ bit-vectors, where $n$ is the bit-width. Each bit-vector is formed by taking each bit position of the quantized values. Figure~\ref{fig:representation}-(b) illustrates the bitplane-based representation. In this representation, any runtime request of reduced bit-width directly translates into proportional speedup, as we can simply load the specified amount of bits. While relatively common in CPU GEMM implementations~\cite{serial1, serial2}, its adoption in GPUs has not gained much attention yet. Concurrent to our work, LUT-GEMM proposes a quantized matrix-vector multiplication kernel adopting a kind of bitplane-based weight representation~\cite{lutgemm}, but it lacks support for any-precision as it necessitates the generation of distinct weight layouts to accommodate different bit-widths. More importantly, LUT-GEMM strictly requires weights to adhere to a specific format called BCQ (binary-coding quantization) --- BCQ cannot support the codebook-based non-uniform quantization method, which we identify as optimal for any-precision LLM. Hence, a novel software engine that fully supports both bitplane-based weight representation and non-uniform quantization is needed. 



\begin{figure}[t]
    \centering
    \includegraphics[width=\linewidth]{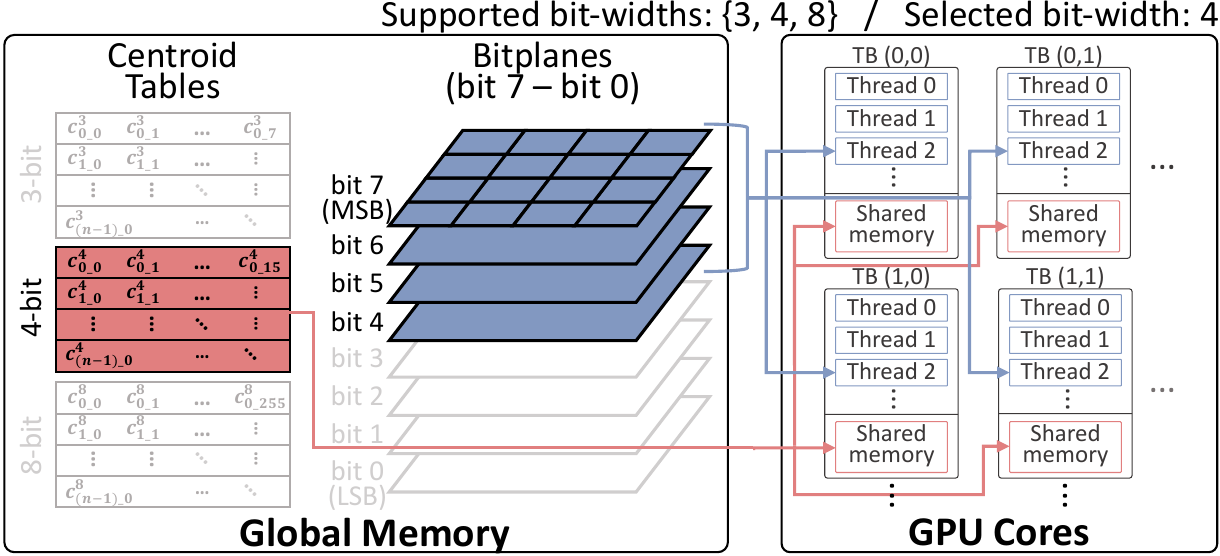}
    \caption{Overview of our specialized engine for any-precision LLM. We illustrate the engine running a 4-bit model while support for 3, 4 and 8-bit is available.}
    \label{fig:overview}
\end{figure}

\begin{figure*}[t]
    \centering
    \includegraphics[width=\textwidth]{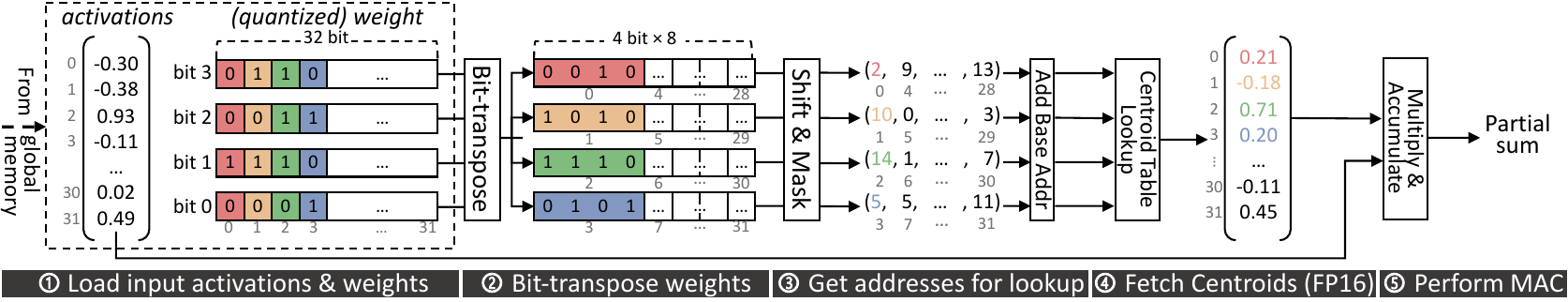}
    \caption{Five steps of thread-level operations.}
    \label{fig:operations}
\end{figure*}

\subsection{System Overview}
\label{sec:system:overview}

Figure~\ref{fig:overview} is an overview of our engine for any-precision LLM. This figure assumes a scenario where support for 3, 4 and 8-bit is available, and the engine is requested to run a 4-bit model. For simplicity, we show only a single linear operation. Weight matrix bitplanes of bit 0 to bit 7, along with tables containing centroid values for the supported bit-widths (3, 4, 8-bit), are stored in memory. Assuming channel-wise quantization, the centroid tables have rows equal to the number of output channels ($n$), with each row containing $2^{k}$ values, where $k$ refers to the bit-width. For each operation, only necessary data is loaded from memory, which in this case includes the centroid table for 4-bit and the bitplanes for bit 4, 5, 6, and 7 in this example. The rows of the centroid table are scattered across the shared memory of different thread blocks since they are frequently accessed by all the threads in the block. Meanwhile, the threads load non-overlapping regions of the weight bitplanes.

\para{Thread-Level Operations}
Figure~\ref{fig:operations} depicts the five step thread-level operation, assuming a bit-width of 4. \circled{1} First, each thread starts by loading 32 input activation values along with their corresponding weights, which are four 32-bit bit-vectors. \circled{2} Next, the four bit-vectors are rearranged so that the bits of each weight align contiguously. This operation is equivalent to the bit-transpose of eight 4-by-4 bit matrices. \circled{3} Subsequently, the bit-transposed bit-vectors are shifted and masked to obtain the indices for centroid lookup. \circled{4} The dequantization process is then completed by fetching centroids from the table in the shared memory. \circled{5} As the final step, Multiply-Accumulate (MAC) operations are conducted on the dequantized weights with the input activations. Threads iteratively execute these five steps until they collectively complete processing the channels assigned to them, alongside other threads in the warp.
 
\subsection{GPU Kernel Optimization}
\label{sec:system:optimization}

\begin{figure}[t]
    \centering
    \includegraphics[width=\linewidth]{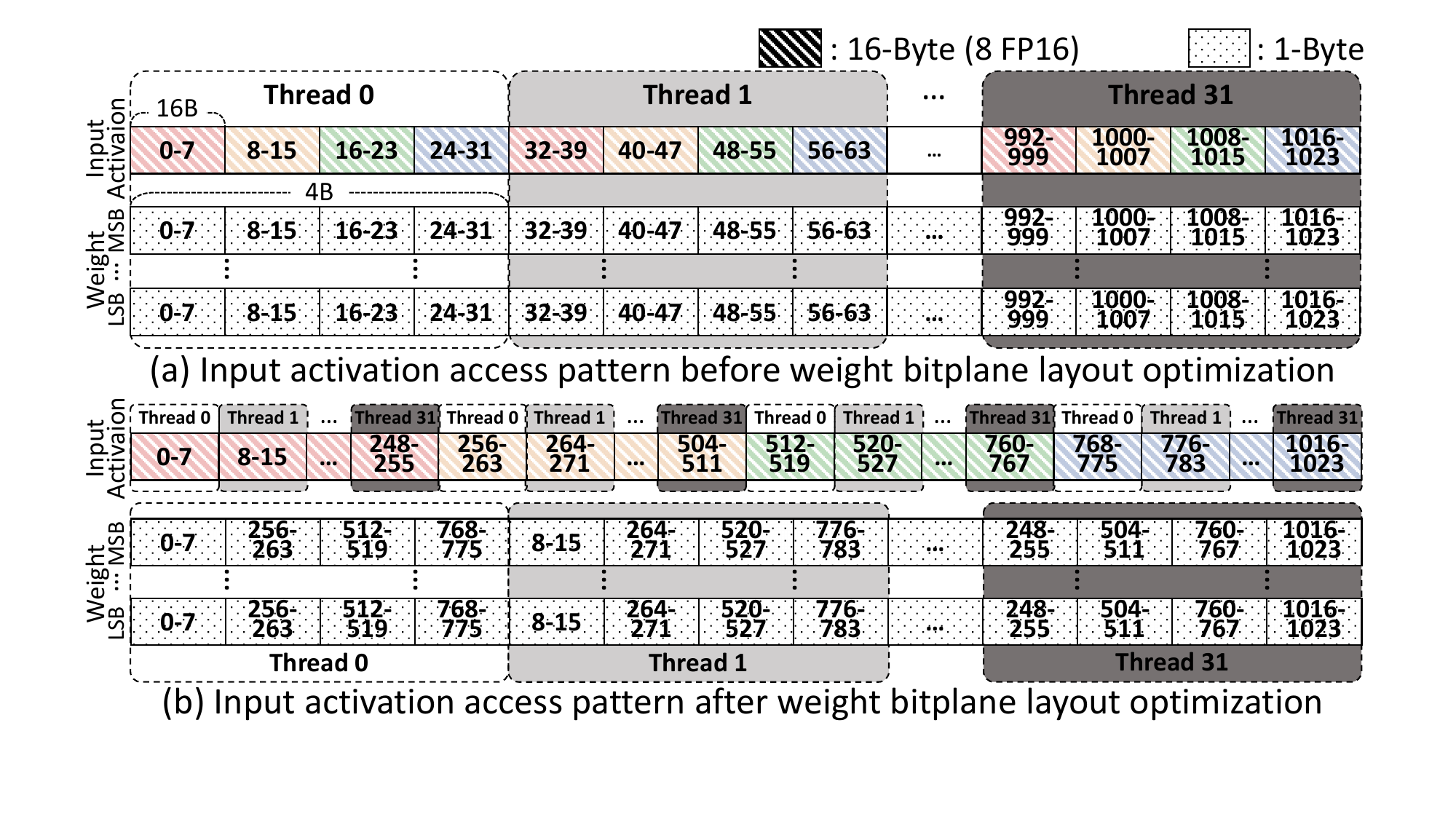}
    \caption{Memory access pattern of the 32 threads in a warp (a) before and (b) after the weight bitplane layout optimization. Simultaneously accessed input activation blocks are in the same colors.}
    \label{fig:permute}
\end{figure}

This section describes three GPU kernel optimization techniques aimed at addressing inefficiencies stemming from the characteristics of bitplane-based quantized GEMM.

\para{Weight Bitplane Layout Optimization}
The bitplane-based representation causes a unique problem for input activation loading. Figure~\ref{fig:permute}-(a) illustrates this issue. When reading from GPU memory, it is preferable for all 32 threads in a warp to access consecutive memory locations so that the memory accesses can be coalesced into a single request. As memory access granularity is 128 bytes, it is optimal when each one of the 32 threads in a warp load consecutive 4-byte blocks (32 weights) from each weight bitplane, to a total of 128 bytes. Furthermore, threads should also load the corresponding consecutive 32 input activation values. Since each activation value is in FP16, each thread needs to load a 64-byte (32$\times$2) block. As CUDA limits the per-thread maximum load size to 16 bytes, four memory accesses are necessary. At each of the four memory accesses, the 32 threads access non-contiguous locations in memory --- for the first access, this is activations 0-7 for thread 0, activations 32-39 for thread 1, and so on --- which is not ideal.

Hence, we suggest permuting bytes in the bitplanes to ensure that threads access activations in a coalesced manner. Figure~\ref{fig:permute}-(b) visualizes the transformation in the weight bitplane layout and the resulting activation access pattern. The indices of weights accessed by each thread are no longer sequential; instead, the indices of weights in each byte of a 4-byte block are now 256 units apart . For instance, thread 0 processes weights 0-7, 256-263, 512-519, and 768-775, as opposed to weights 0-31. Consequently, each memory access of the 32 threads for the activations leads to a coalesced memory access pattern. Such byte permutation of bitplanes is performed as a part of pre-processing.

\begin{figure}[t]
    \centering
    \includegraphics[width=\linewidth]{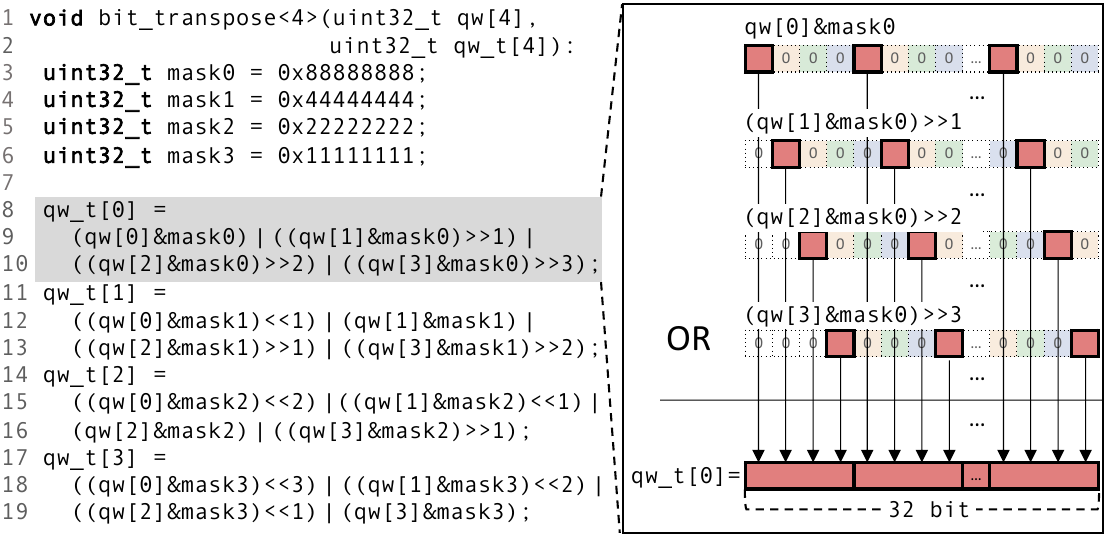}
    \caption{Efficient bit-transpose operation of our engine.}
    \label{fig:transpose}
\end{figure}

\para{Efficient Bit-transpose}
In order to dequantize weights, we have to preform a bit-transpose step (step 2 in Figure~\ref{fig:operations}) as the weights are represented as bitplanes. This step requires a large number of bitwise operations, becoming the main overhead. Even an optimized bit-transpose algorithm requires 38 bitwise operations just to process an 8$\times$8 bit matrix~\cite{hacker}. 

Figure~\ref{fig:transpose} describes our optimized bit-transpose algorithm. The example in the figure assumes a bit-width of 4, taking four 32-bit bit-vectors as input (\texttt{qw[4]}) and outputting four new 32-bit bit-vectors (\texttt{qw\_t[4]}) in which the bits of weights are arranged contiguously. The key is to treat a 32-bit bit-vector as eight 4-bit sub-bit-vectors, ensuring that operations on them effectively function like SIMD (Single Instruction Multiple Data) operations. With this approach, we can pack bit positions 0, 4, ..., and 28 from the four input bit-vectors into a single 32-bit bit-vector with 10 bitwise operations. This process is repeated three more times for the other bit positions, completing the entire task. This totals to 40 bitwise operations --- nearly half the number required (76) if you were to apply the 8$\times$8 bit matrix transpose algorithm twice instead. This approach is also applicable for bit-widths of 2 and 8-bit by interpreting a 32-bit bit-vector as sixteen 2-bit and four 8-bit sub-bit-vectors, respectively. For non-power-of-2 bit-widths, we seamlessly apply the algorithm for the next larger bit-width that is a power of 2.

\begin{figure}[t]
    \centering
    \includegraphics[width=\linewidth]{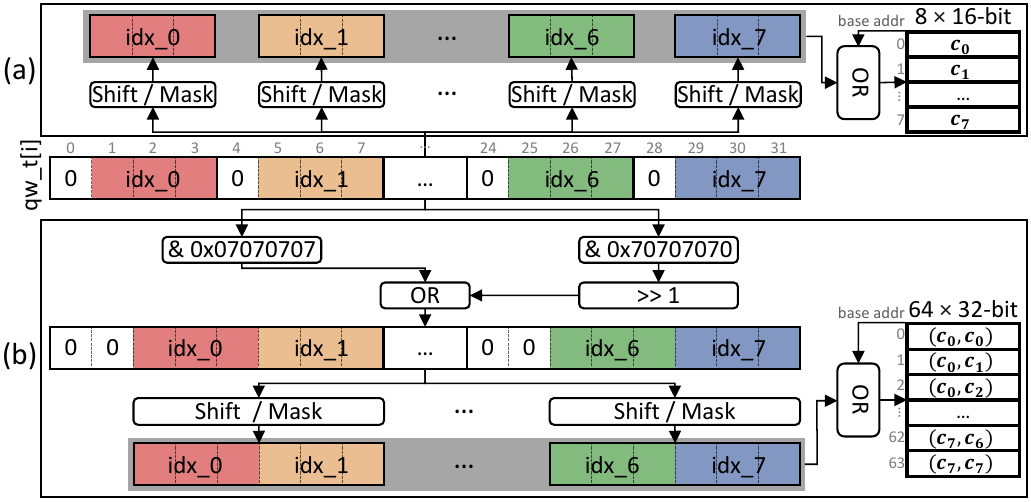}
    \vspace{-0.3cm}
    \caption{Operations to obtain centroid table lookup addresses (a) before and (b) after table lookup merging.}
    \vspace{-0.2cm}
    \label{fig:pairwise}
\end{figure}
\para{Merging Table Lookups}
Bitwise operations for the centroid table index calculation (Step 3 in Figure~\ref{fig:operations}) can also become a bottleneck, particularly at small bit-widths. Figure~\ref{fig:pairwise}-(a) shows how this step works in detail, assuming a bit-width of 3. After Step 2 of Figure~\ref{fig:operations}, we obtain four 32-bit bit-vectors (\texttt{qw\_t[0]},~$...$,~\texttt{qw\_t[3]}), each containing eight 3-bit indices (\texttt{idx\_0},~$...$,~\texttt{idx\_7}) zero-extended to 4 bits. To complete the dequantization for each bit-vector, eight centroid table lookups are necessary, and each lookup entails three bitwise operations: the first two for the shifting and masking to derive indices, and the third for adding the base address of the table to the indices. Thus, 24 bitwise operations are required for each bit-vector.  

To alleviate this overhead on the 3-bit case, we halve the number of table lookups by merging two lookups into one, as depicted in Figure~\ref{fig:pairwise}-(b). Two adjacent 3-bit indices ([\texttt{idx\_0}, \texttt{idx\_1}], [\texttt{idx\_2}, \texttt{idx\_3}],~$...$) are merged into a single 6-bit value, serving as a new index for the lookup. The table is expanded to contain 64 entries instead of 8, containing all possible pairs of centroids ([$c_0$, $c_0$], [$c_0$, $c_1$], $...$). Now we retrieve two centroid values with a single lookup. The number of required bitwise operations is reduced to 16 for each 32-bit bit-vector: 12 for shifting, masking, and adding the base address, and the remaining 4 for rearranging the 32-bit bit-vector. By allowing an acceptable increase in shared memory usage, we mitigate the compute costs.
\section{Evaluation}
\label{sec:eval}
Through extensive experiments, we demonstrate that our proposal is an effective way of deploying multiple, different-sized LLMs by proving the following two arguments:
\vspace{-0.3cm}
\begin{itemizeReduced}
    \item A set of quantized models generated from incremental upscaling, when integrated with the non-unifrom quantization method, match the state-of-the-art quantization results at their respective bit-widths. (Section~\ref{sec:eval:accuracy})
    \item Our specialized engine matches or even outperforms existing engines while providing memory-efficient any-precision support, a feature lacking in the existing engines. (Section~\ref{sec:eval:microbenchmarks} \& Section~\ref{sec:eval:end-to-end})
\end{itemizeReduced}

\subsection{Any-Precision Quantization Results}
\label{sec:eval:accuracy}

\para{Methodology} We evaluate 4 to 8-bit models obtained through incremental upscaling, using a 3-bit SqueezeLLM model as the seed model. The results are compared with the 4 to 8-bit SqueezeLLM models.
We benchmark our method on LLaMA-2-7B~\cite{llama2}, Mistral-7B~\cite{mistral}, and three OPT models (6.7B, 2.7B, 1.3B)~\cite{opt}. We evaluate the models with two metrics: perplexity on three datasets (WikiText2~\cite{wikitext}, PTB~\cite{ptb}, C4~\cite{c4}) and zero-shot accuracy on five tasks (ARC-easy/challenge~\cite{arc}, HellaSwag~\cite{hellaswag}, PIQA~\cite{piqa}, WinoGrande~\cite{winogrande}). Experimental details are in Appendix~\ref{appendix:settings}.

\begin{table*}[t]
 \caption{Perplexity on Wikitext2 (Wiki), C4 and Penn Treebank (PTB) for vanilla SqueezeLLM (SqLLM) and SqueezeLLM integrated with incremental upscaling (SqLLM+IU) using a 3-bit seed model. We also report the difference between the two methods, highlighting cases in red where the increase in perplexity exceeds 0.1.}
 \label{table:perplexity}
 \resizebox{\textwidth}{!}{%
 \begin{tabular}{cc|ccc|ccc|ccc|ccc|ccc|ccc|c|ccc|}

& & \multicolumn{3}{c}{3-bit} & \multicolumn{3}{c}{4-bit} & \multicolumn{3}{c}{5-bit} & \multicolumn{3}{c}{6-bit} & \multicolumn{3}{c}{7-bit} & \multicolumn{3}{c|}{8-bit} & & \multicolumn{3}{c|}{FP16} \\
& & Wiki & PTB & C4 & Wiki & PTB & C4 & Wiki & PTB & C4 & Wiki & PTB & C4 & Wiki & PTB & C4 & Wiki & PTB & C4 & & Wiki & PTB & C4 \\ \cline{1-20} \cline{22-24}
 \multicolumn{1}{c}{\multirow{3}{*}{\textbf{Llama-2-7B}}} & SqLLM & 6.13 & 8.95 & 8.20 & 5.61 & 8.24 & 7.44 & 5.50 & 8.13 & 7.30 & 5.47 & 8.11 & 7.27 & 5.47 & 8.10 & 7.27 & 5.47 & 8.10 & 7.26 & & \multirow{3}{*}{5.47} & \multirow{3}{*}{8.10} & \multirow{3}{*}{7.26} \\
 \multicolumn{1}{c}{} & SqLLM+IU & - & - & - & 5.62 & 8.23 & 7.45 & 5.50 & 8.12 & 7.30 & 5.47 & 8.10 & 7.27 & 5.47 & 8.10 & 7.27 & 5.47 & 8.10 & 7.26 & & & & \\
 \multicolumn{1}{c}{} & $\Delta$ & - & - & - & $\blacktriangle$ 0.01 & $\blacktriangledown$ 0.01 & $\blacktriangle$ 0.01 & - & $\blacktriangledown$ 0.01 & - & - & $\blacktriangledown$ 0.01 & - & - & - & - & - & - & - & & & & \\
 \cline{1-20} \cline{22-24}
 \multicolumn{1}{c}{\multirow{3}{*}{\textbf{Mistral-7B}}} & SqLLM & 5.94 & 9.61 & 9.33 & 5.36 & 8.65 & 8.55 & 5.27 & 8.52 & 8.41 & 5.25 & 8.47 & 8.39 & 5.25 & 8.46 & 8.38 & 5.25 & 8.46 & 8.38 & & \multirow{3}{*}{5.25} & \multirow{3}{*}{8.46} & \multirow{3}{*}{8.38} \\
 \multicolumn{1}{c}{} & SqLLM+IU & - & - & - & 5.39 & 8.68 & 8.57 & 5.28 & 8.52 & 8.42 & 5.25 & 8.48 & 8.39 & 5.25 & 8.46 & 8.38 & 5.24 & 8.46 & 8.38 & & & & \\
 \multicolumn{1}{c}{} & $\Delta$ & - & - & - & $\blacktriangle$ 0.03 & $\blacktriangle$ 0.03 & $\blacktriangle$ 0.02 & $\blacktriangle$ 0.01 & - & $\blacktriangle$ 0.01 & - & $\blacktriangle$ 0.01 & - & - & - & - & $\blacktriangledown$ 0.01 & - & - & & & & \\
 \cline{1-20} \cline{22-24}
 \multicolumn{1}{c}{\multirow{3}{*}{\textbf{OPT-6.7B}}} & SqLLM & 11.60 & 13.35 & 13.83 & 10.96 & 12.59 & 13.17 & 10.84 & 12.55 & 13.07 & 10.84 & 12.57 & 13.05 & 10.86 & 12.52 & 13.05 & 10.86 & 12.52 & 13.05 & & \multirow{3}{*}{10.86} & \multirow{3}{*}{12.52} & \multirow{3}{*}{13.05} \\
 \multicolumn{1}{c}{} & SqLLM+IU & - & - & - & 11.01 & 12.82 & 13.20 & 10.88 & 12.55 & 13.07 & 10.84 & 12.55 & 13.05 & 10.84 & 12.55 & 13.05 & 10.86 & 12.55 & 13.05 & & & & \\
 \multicolumn{1}{c}{} & $\Delta$ & - & - & - & $\blacktriangle$ 0.05 & {\color{red}$\blacktriangle$} 0.23 & $\blacktriangle$ 0.03 & $\blacktriangle$ 0.04 & - & - & - & $\blacktriangledown$ 0.02 & - & $\blacktriangledown$ 0.02 & $\blacktriangle$ 0.03 & - & - & $\blacktriangle$ 0.03 & - & & & & \\
 \cline{1-20} \cline{22-24}
 \multicolumn{1}{c}{\multirow{3}{*}{\textbf{OPT-2.7B}}} & SqLLM & 13.97 & 15.86 & 16.11 & 12.70 & 14.75 & 14.95 & 12.42 & 14.47 & 14.78 & 12.47 & 14.41 & 14.75 & 12.45 & 14.41 & 14.73 & 12.47 & 14.44 & 14.73 & & \multirow{3}{*}{12.47} & \multirow{3}{*}{14.44} & \multirow{3}{*}{14.73} \\
 \multicolumn{1}{c}{} & SqLLM+IU & - & - & - & 12.72 & 14.78 & 15.02 & 12.47 & 14.52 & 14.78 & 12.45 & 14.47 & 14.75 & 12.47 & 14.44 & 14.75 & 12.47 & 14.44 & 14.73 & & & & \\
 \multicolumn{1}{c}{} & $\Delta$ & - & - & - & $\blacktriangle$ 0.02 & $\blacktriangle$ 0.03 & $\blacktriangle$ 0.07 & $\blacktriangle$ 0.05 & $\blacktriangle$ 0.05 & - & $\blacktriangledown$ 0.02 & $\blacktriangle$ 0.06 & - & $\blacktriangle$ 0.02 & $\blacktriangle$ 0.03 & $\blacktriangle$ 0.02 & - & - & - & & & & \\
 \cline{1-20} \cline{22-24}
 \multicolumn{1}{c}{\multirow{3}{*}{\textbf{OPT-1.3B}}} & SqLLM & 16.30 & 18.39 & 18.39 & 14.93 & 16.45 & 16.81 & 14.64 & 16.20 & 16.62 & 14.61 & 16.20 & 16.55 & 14.61 & 16.20 & 16.55 & 14.61 & 16.17 & 16.55 & & \multirow{3}{*}{14.64} & \multirow{3}{*}{16.17} & \multirow{3}{*}{16.55} \\
 \multicolumn{1}{c}{} & SqLLM+IU & - & - & - & 14.95 & 16.59 & 16.88 & 14.66 & 16.23 & 16.62 & 14.64 & 16.20 & 16.55 & 14.64 & 16.20 & 16.55 & 14.64 & 16.17 & 16.55 & & & & \\
 \multicolumn{1}{c}{} & $\Delta$ & - & - & - & $\blacktriangle$ 0.02 & {\color{red}$\blacktriangle$} 0.14 & $\blacktriangle$ 0.07 & $\blacktriangle$ 0.02 & $\blacktriangle$ 0.03 & - & $\blacktriangle$ 0.03 & - & - & $\blacktriangle$ 0.03 & - & - & $\blacktriangle$ 0.03 & - & - & & & & \\
 \cline{1-20} \cline{22-24}

 \end{tabular}%
 }
\end{table*}

\begin{table}[t]
    \caption{Average zero-shot accuracy on five tasks --- ARC-easy, ARC-challenge, HellaSwag, PIQA, and WinoGrande --- for vanilla SqueezeLLM (SqLLM) and SqueezeLLM integrated with incremental upscaling (SqLLM+IU) using a 3-bit seed model. We also report the difference between the two methods.}
    \label{table:zero-shot-avg}
\resizebox{\linewidth}{!}{%
\begin{tabular}{cccccccc|l|c}
                            &          & 3-bit & 4-bit                   & 5-bit                   & 6-bit                   & 7-bit               & 8-bit                   &  & FP16                  \\ \cline{1-8} \cline{10-10} 
\multirow{3}{*}{Llama-2-7B} & SqLLM    & 66.2  & 68.3                    & 68.6                    & 68.8                    & 68.9                & 68.9                    &  & \multirow{3}{*}{69.0} \\
                            & SqLLM+IU & -     & 68.2                    & 68.8                    & 68.9                    & 68.9                & 69.0                    &  &                       \\
                            & $\Delta$ & -     & $\blacktriangledown$0.1 & $\blacktriangle$0.2     & $\blacktriangle$0.1     & -                   & $\blacktriangle$0.1     &  &                       \\ \cline{1-8} \cline{10-10} 
\multirow{3}{*}{Mistral-7B} & SqLLM    & 71.5  & 73.2                    & 73.8                    & 74.0                    & 74.1                & 74.2                    &  & \multirow{3}{*}{74.1} \\
                            & SqLLM+IU & -     & 73.3                    & 73.8                    & 74.0                    & 74.2                & 74.1                    &  &                       \\
                            & $\Delta$ & -     & $\blacktriangle$0.1     & -                       & -                       & $\blacktriangle$0.1 & $\blacktriangledown$0.1 &  &                       \\ \cline{1-8} \cline{10-10} 
\multirow{3}{*}{OPT-6.7B}   & SqLLM    & 58.6  & 59.9                    & 60.6                    & 60.7                    & 60.7                & 60.8                    &  & \multirow{3}{*}{60.8} \\
                            & SqLLM+IU & -     & 60.3                    & 60.6                    & 60.7                    & 60.8                & 60.7                    &  &                       \\
                            & $\Delta$ & -     & $\blacktriangle$0.4     & -                       & -                       & $\blacktriangle$0.1 & $\blacktriangledown$0.1 &  &                       \\ \cline{1-8} \cline{10-10} 
\multirow{3}{*}{OPT-2.7B}   & SqLLM    & 54.3  & 55.8                    & 56.3                    & 56.2                    & 56.3                & 56.3                    &  & \multirow{3}{*}{56.4} \\
                            & SqLLM+IU & -     & 55.7                    & 56.1                    & 56.2                    & 56.4                & 56.3                    &  &                       \\
                            & $\Delta$ & -     & $\blacktriangledown$0.1 & $\blacktriangledown$0.2 & -                       & $\blacktriangle$0.1 & -                       &  &                       \\ \cline{1-8} \cline{10-10} 
\multirow{3}{*}{OPT-1.3B}   & SqLLM    & 50.7  & 52.7                    & 52.9                    & 53.4                    & 53.4                & 53.3                    &  & \multirow{3}{*}{53.3} \\
                            & SqLLM+IU & -     & 52.6                    & 53.1                    & 53.3                    & 53.4                & 53.3                    &  &                       \\
                            & $\Delta$ & -     & $\blacktriangledown$0.1 & $\blacktriangle$0.2     & $\blacktriangledown$0.1 & -                   & -                       &  &                       \\ \cline{1-8} \cline{10-10} 
\end{tabular}%
}
\end{table}

\para{Results} 
Table~\ref{table:perplexity} shows the perplexity results. The upscaled models (SqLLM+IU) nearly match the independently quantized models (SqLLM) across various models, datasets, and bit-widths. Except in two cases (OPT-6.7B, OPT-1.3B 4-bit on PTB), the increase in perplexity is negligible (< 0.1). 

A similar trend is observed in the zero-shot task results. Table~\ref{table:zero-shot-avg} presents the average accuracy across five zero-shot tasks. The upscaled models achieve the same level of accuracy as the independently quantized models. The accuracy drop from upscaling is within 0.2\%, with upscaled models even outperforming in some cases. Full results for each task are provided in Appendix~\ref{appendix:additional-quant}.

These results demonstrate that with a single $n$-bit parent model generated through incremental upscaling, we can utilize the full range of 3 to $n$-bit models, all achieving state-of-the-art quality at their respective bit-widths.

\begin{table}[t]
\caption{Runtime (sec) of the proposed any-precision quantization scheme, composed of two stages: seed model generation (Seed Gen) and incremental upscaling (IU).}
\label{table:runtime}
\resizebox{\linewidth}{!}{%
\begin{tabular}{cccccc}
                      & Llama-2-7B & Mistral-7B & OPT-6.7B & OPT-2.7B & OPT-1.3B \\ \hline
Seed Gen  &  36.2 &  37.2  & 37.0 & 14.0  & 6.4  \\
IU & 15.6   & 18.2  & 12.8   & 6.2    & 3.0    \\ \hline
Total                 &   51.8         &    55.4        &    49.8      &    20.2      &   9.4      
\end{tabular}
}
\end{table}

\para{Runtime of Any-Precision Quantization} 
Our any-precision quantization scheme is efficient as it does not require training and is highly parallelizable. We measure the runtime of the any-precision quantization process, beginning with a 3-bit seed model and progressing up to the final 8-bit parent model, on an Intel i9-13900K CPU with 24 cores. Table~\ref{table:runtime} shows the results. It takes less than a minute to complete the whole process, even for 7B-scale models.

\subsection{Kernel Microbenchmarks}
\label{sec:eval:microbenchmarks}

\begin{table*}[t]
\caption{Matrix-vector multiplication speedup of our kernel over the cuBLAS FP16 baseline on three weight sizes of Llama-2-7B, compared against the existing kernel for non-uniform quantization (SqLLM).}
\label{table:microbenchmarks}
    \centering
    \resizebox{\textwidth}{!}{
    \begin{tabular}{c|c|cccccc|cccccc|cccccc|}
         & & \multicolumn{6}{c|}{(1, 4096) $\times$ (4096, 4096)} & \multicolumn{6}{c|}{(1, 4096) $\times$ (11008, 4096)} & \multicolumn{6}{c|}{(1, 11008) $\times$ (4096, 11008)} \\
         & & 3-bit & 4-bit & 5-bit & 6-bit & 7-bit & 8-bit & 3-bit & 4-bit & 5-bit & 6-bit & 7-bit & 8-bit & 3-bit & 4-bit & 5-bit & 6-bit & 7-bit & 8-bit \\ \hline
        \multicolumn{1}{c|}{\multirow{2}{*}{\textbf{RTX 4090}}} & Ours & $\times$3.99 & $\times$3.03 & $\times$2.61 & $\times$2.18 & $\times$1.87 & $\times$1.56 & $\times$4.67 & $\times$3.43 & $\times$2.82 & $\times$2.44 & $\times$2.13 & $\times$1.78 & $\times$4.45 & $\times$3.42 & $\times$2.74 & $\times$2.28 & $\times$2.08 & $\times$1.81 \\
         & SqLLM & $\times$3.69 & $\times$3.07 & - & - & - & - & $\times$4.08 & $\times$3.12 & - & - & - & - & $\times$4.22 & $\times$3.17 & - & - & - & - \\ \hline
        \multicolumn{1}{c|}{\multirow{2}{*}{\textbf{\begin{tabular}[c]{@{}c@{}}RTX 4070 \\ Laptop\end{tabular}}}} & Ours & $\times$4.97 & $\times$3.73 & $\times$3.01 & $\times$2.51 & $\times$2.10 & $\times$1.76 & $\times$5.15 & $\times$3.84 & $\times$3.07 & $\times$2.46 & $\times$2.06 & $\times$1.72 & $\times$5.29 & $\times$3.66 & $\times$3.05 & $\times$2.52 & $\times$2.13 & $\times$1.87 \\
         & SqLLM & $\times$4.74 & $\times$3.66 & - & - & - & - & $\times$5.05 & $\times$3.76 & - & - & - & - & $\times$5.29 & $\times$3.93 & - & - & - & - \\ \hline
        \multicolumn{1}{c|}{\multirow{2}{*}{\textbf{\begin{tabular}[c]{@{}c@{}}Jetson \\ AGX Orin\end{tabular}}}} & Ours & $\times$3.84 & $\times$3.02 & $\times$2.56 & $\times$2.33 & $\times$2.10 & $\times$1.78 & $\times$4.20 & $\times$3.31 & $\times$2.72 & $\times$2.47 & $\times$2.18 & $\times$1.84 & $\times$4.35 & $\times$2.96 & $\times$2.54 & $\times$2.52 & $\times$2.16 & $\times$1.86 \\
         & SqLLM & $\times$3.15 & $\times$1.94 & - & - & - & - & $\times$3.30 & $\times$1.98 & - & - & - & - & $\times$3.36 & $\times$2.04 & - & - & - & - \\ \hline
    \end{tabular}%
    }
\end{table*}

\para{Methodology} We evaluate the latency of matrix-vector multiplication of our kernel on the three matrix dimensions used in Llama-2-7B, and compare with the existing kernel for non-uniform quantization proposed in SqueezeLLM~\cite{sqllm}. Note that the SqueezeLLM kernel only supports 3 and 4 bits. We conduct experiments on three GPUs of varying scales: RTX 4090 (desktop), RTX 4070 Laptop (laptop), and Jetson AGX Orin 64 GB (mobile). We use NVIDIA Nsight Compute to measure latency.

\para{Results}
Table~\ref{table:microbenchmarks} shows the results. Our kernel consistently achieves low latency for various matrix sizes and platforms, showing a near-linear improvement against the cuBLAS FP16 baseline as the bit-width decreases. Compared to the SqueezeLLM kernel, our kernel performs on par on the RTX 4090 and RTX 4070 Laptop, and goes on to exhibit substantial improvements on Jetson. Note that our kernel achieves the performance despite adopting the bitplane-based weight representation to support any-precision, which entails non-trivial engineering challenges such as addressing the increased amount of bitwise operations for bit-transpose. Additional results with a wider range of matrix dimensions and comparison against uniform quantization kernels can be found in Appendix~\ref{appendix:additional-microbenchmarks:gemv} and ~\ref{appendix:additional-microbenchmarks:kernels}.

\begin{figure}[t]
    \centering
    \includegraphics[width=\linewidth]{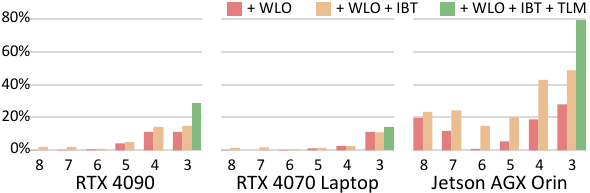}
    \caption{Ablation study of kernel optimization techniques. The speedup over the baseline for matrix-vector multiplication of dimensions (1,4096) and (11008,4096) is reported.}
    \label{fig:ablation}
\end{figure}

\para{Ablation Study} Figure~\ref{fig:ablation} shows how the kernel performance improves with the introduction of each of the three optimization techniques: 1) weight bitplane layout optimization (WLO); 2) improved bit-transpose algorithm over the existing one~\cite{hacker} (IBT); and 3) table lookup merging (TLM) for the case of 3-bit. While all three optimization techniques significantly contribute to improving kernel performance, there are two notable trends. First, the optimization effect is more pronounced on lower bit-widths. This is because, at higher bit-widths, the global memory bandwidth tends to be the sole bottleneck, whereas our techniques primarily optimize cache access patterns (WLO) and computations (IBT, TLM) --- aspects that become more critical at lower bit-widths. Second, the optimization effect is more pronounced on Jetson AGX Orin. This is attributed to its limited on-chip cache and compute resources.

\begin{figure}[t]
    \centering
    \includegraphics[width=\linewidth]{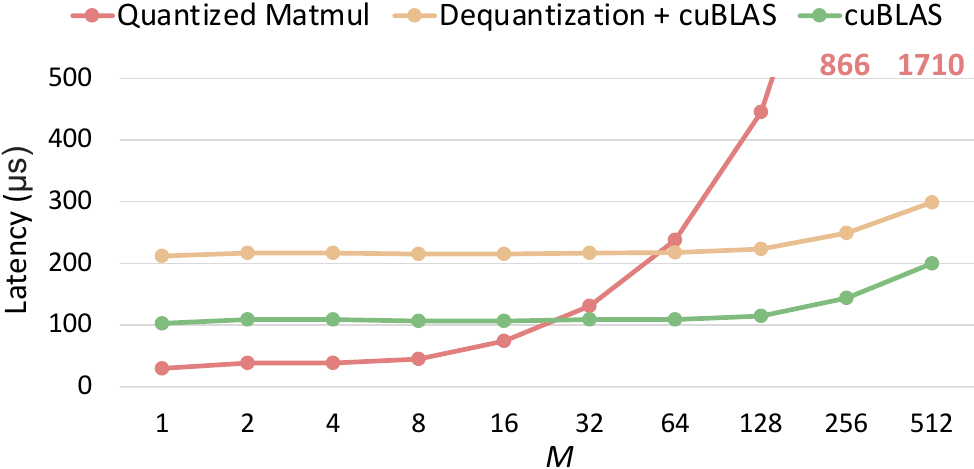}
    \caption{Latency comparison of 4-bit quantized matrix-matrix multiplication using our kernel, separate dequantization followed by cuBLAS FP16 kernel, and cuBLAS FP16 kernel on RTX 4090. Matrix dimensions are ($M$, 4096) and (11008, 4096), with $M$ varying from 1 to 512.}
    \label{fig:matmul}
\end{figure}

\para{Matrix-Matrix Multiplication Performance} Even for our target setting of running a model on personal devices, where a single query is typically served at a time, efficient matrix-matrix multiplication is necessary for two reasons. First, the batch size might not always be 1; it can be small (e.g., 2, 4, 8). Even in a single query, multiple tokens are generated in parallel when using advanced decoding algorithms like beam search~\cite{seq-to-seq} and parallel sampling. Speculative decoding also requires the parallel processing of multiple tokens. Second, during the prefill phase, the tokens in the input prompt are processed together, with their number potentially reaching into the hundreds or thousands.

The red line in Figure~\ref{fig:matmul} illustrates the latency of our kernel for 4-bit quantized matrix-matrix multiplication with dimensions ($M$, 4096) and (11008, 4096) on an RTX 4090, with $M$ varying from 1 to 512. Our kernel performs robustly with small $M$ values such as 2, 4, and 8, showing a significant latency gap compared to the cuBLAS baseline (green line). This indicates that our kernel is also effective for generation with small batch sizes. Similar results are observed for other bit-widths and platforms, as detailed in Appendix~\ref{appendix:additional-microbenchmarks:gemm}.

When $M$ becomes very large, however, as in the prefill phase, our kernel becomes slower than the cuBLAS FP16 baseline (green line). This slowdown occurs because our kernel does not use tensor cores, making it more susceptible to increases in computation with large $M$. To mitigate potential slowdown of the prefill phase, our engine dequantizes the weights using a separate kernel and then employs the cuBLAS kernel when $M$ exceeds a certain threshold (e.g., 16). The latency gap between the dequantization + cuBLAS (represented by the yellow line) and cuBLAS remains almost constant (about 100 µs), as dequantization time does not scale with $M$. This limited performance degradation of the prefill phase is small enough to be easily amortized when the generation length is sufficiently long. 


\begin{figure}[t]
    \centering
    \includegraphics[width=\linewidth]{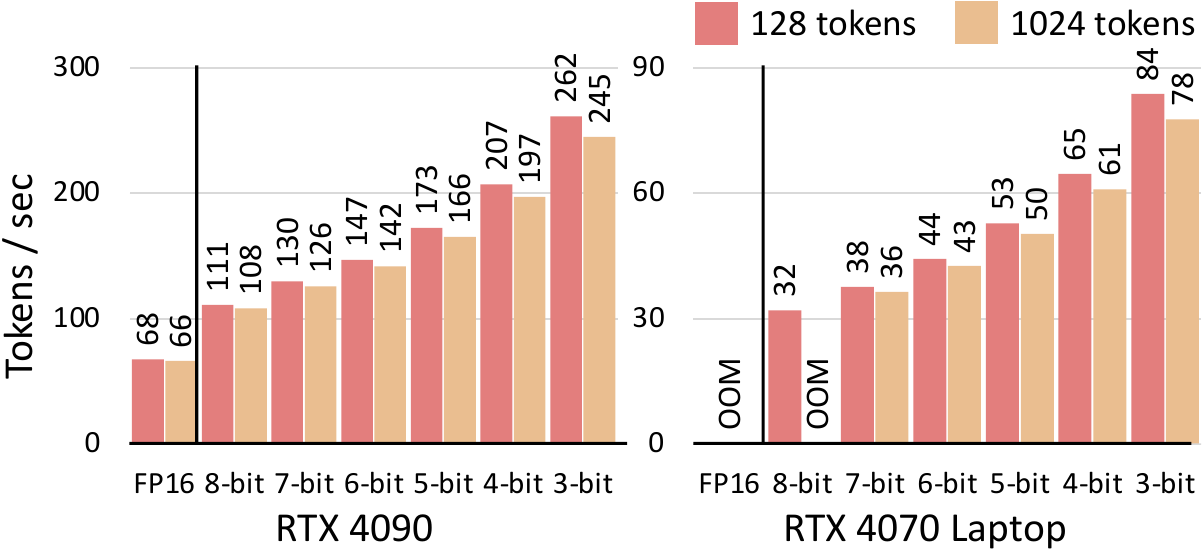}
    \caption{End-to-end throughput of Llama-2-7B.}
    \label{fig:end-to-end}
\end{figure}

\subsection{End-to-end Throughput}
\label{sec:eval:end-to-end}
We evaluate the end-to-end inference throughput of our engine by integrating it with TensorRT-LLM~\cite{tensorrt-llm}. Figure ~\ref{fig:end-to-end} shows the throughput of Llama-2-7B in generating 128 and 1024 tokens on RTX 4090 and RTX 4070 Laptop. TensorRT-LLM currently lacks  support for Jetson. The enhanced kernel performance effectively translates into an end-to-end speedup. A slightly lower speedup with a longer sequence length (1024) is due to the increased overhead of attention, which is not accelerated by weight quantization. Results for the other models are in Appendix~\ref{appendix:additional-end-to-end}.


\section{Conclusion}
\label{sec:conclusion}
We make a case for any-precision LLM, which enables memory-efficient and cost-effective deployment of multiple, different-sized LLMs. We propose a lightweight method for any-precision quantization of LLM, along with a specialized software engine to fully leverage its benefits.

\section*{Impact Statement}
This paper presents work whose goal is to advance the field of Machine Learning. There are many potential societal consequences of our work, none of which we feel must be specifically highlighted here.

\section*{Acknowledgement}
This work was supported by the National Research Foundation of Korea (NRF) grant funded by the Korea Government (MSIT) (RS-2024-00340008), Institute of Information \& Communications Technology Planning \& Evaluation (IITP) funded by the Korea Government (MSIT) (2021-0-00105, Development of Model Compression Framework for Scalable On-device AI Computing on Edge Applications), and Institute of Information \& Communications Technology Planning \& Evaluation (IITP) under the artificial intelligence semiconductor support program (IITP-2023-RS-2023-00256081), funded by the Korea Government (MSIT).


\bibliography{example_paper}
\bibliographystyle{icml2024}

\newpage
\appendix

\section{Evaluation Details}
\label{appendix:settings}

\subsection{Datasets}
\label{appendix:settings:datasets}
For WikiText-2, we concatenate on the test set to form a continuous string for perplexity evaluation. For C4, we concatenate samples from the validation set, as using the whole unsampled dataset is infeasible and impractical due to the large size of the dataset. For the Penn Treebank dataset, we concatenate on the test set. In its original state, rare words are  replaced with "\verb|<unk>|" --- we modify this by splitting each "\verb|<unk>|" into five separate characters: "\verb|<|", "\verb|u|", "\verb|n|", "\verb|k|", and "\verb|>|". This approach prevents the tokenizer from recognizing "\verb|<unk>|" as a special unknown token, which would significantly increase model perplexity due to the inability to predict occurrences of unknown words. By treating each character of "\verb|<unk>|" as a regular token, we maintain consistent token processing.

\subsection{Perplexity Calculations}
\label{appendix:settings:ppl}
To evaluate a given string, we start by tokenizing it with the default HuggingFace tokenizer of each model. Then we chunk the sequence into non-overlapping segments of length 2048, as in the manner of previous works \cite{gptq} \cite{sqllm}. We process these segments through the model to collect the log-probabilities of subsequent token generation. The final perplexity we report is the exponentiated average of these log-probabilities.

\subsection{Zero-shot Tasks}
\label{appendix:settings:zero-shot}

We use the Language Model Evaluation Harness framework \cite{lm-eval} to evaluate our models on zero-shot tasks. We report the byte-length normalized accuracy where applicable.




\section{Full Results on Zero-shot Tasks}
\label{appendix:additional-quant}

The data in Table~\ref{table:zero-shot} presents the full results for each of the five zero-shot tasks (ARC-easy, ARC-challenge, HellaSwag, PIQA, WinoGrande). The results indicate that the SqLLM models with incremental upscaling (SqLLM+IU) match the independently quantized SqLLM models across a range of tasks, models, and bit-widths. 



\section{Additional Kernel Microbenchmark Results}
\label{appendix:additional-microbenchmarks}

\subsection{Kernel Latency on Various Matrix Sizes}
\label{appendix:additional-microbenchmarks:gemv}
Table~\ref{table:various-sizes} presents the latency of of our kernel for matrix-vector multiplication across various matrix sizes. We select the matrix sizes that are used in Llama-2-7B, Mistral-7B, OPT-6.7B, OPT-2.7B and OPT-1.3B.


\subsection{Comparison with Kernels for Uniform Quantization}
\label{appendix:additional-microbenchmarks:kernels}
Table~\ref{table:compare-uniform} provides a comparison of our kernel's matrix-vector multiplication performance against existing kernels designed for uniform quantization: ExLlamaV2~\cite{exllamav2}, LUT-GEMM~\cite{lutgemm}, AWQ~\cite{awq}, TensorRT-LLM~\cite{tensorrt-llm}. Note that this is not an apples-to-apples comparison as they lack any-precision support. They are specifically tailored for uniform quantization and/or employ a bitpacking-based weight representation. Nevertheless, our kernel demonstrates competitive performance, achieving the best results in the majority of cases.

\subsection{Matrix-Matrix Multiplication Performance}
\label{appendix:additional-microbenchmarks:gemm}
Table~\ref{table:multi-batch} shows the matrix-matrix multiplication performance with small batch sizes (2, 4, 8) on our kernel to test its applicability on the small-batch use cases mentioned above. Our kernel demonstrates impressive performance for multiple batches on both the RTX 4090 and RTX 4070 Laptop, with less than a 30\% drop in performance compared to single-batch processing in most scenarios. The exceptions are in cases with very low bit-widths, like 3 and 4-bit, where the performance difference might be more pronounced. This trend slightly varies for Jetson AGX Orin, which has highly limited compute resources, but still delivers commendable results, particularly for batch sizes of 2 and 4.


\section{Additional End-to-End Throughput Evaluation Results}
\label{appendix:additional-end-to-end}

\begin{figure}[t]
    \centering
    \includegraphics[width=\linewidth]{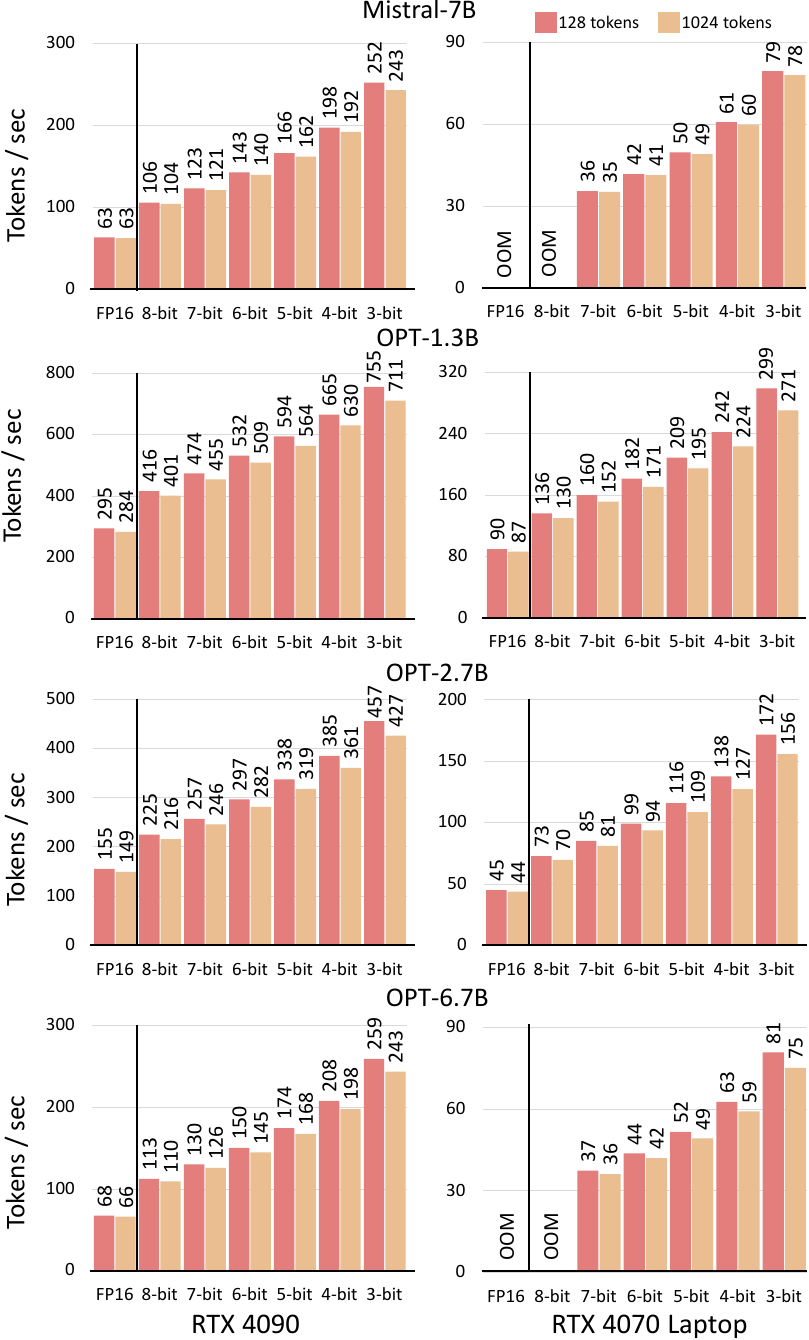}
    \caption{End-to-end throughput of Mistral-7B, OPT-6.7B, OPT-2.7B and OPT-1.3B.}
    \label{fig:end-to-end_appendix}
\end{figure}

Figure~\ref{fig:end-to-end_appendix} shows the end-to-end inference throughput of Mistral-7B, OPT-6.7B, OPT-2.7B and OPT-1.3B.

\section{Incremental Upscaling with Uniform Quantization Methods}
\label{appendix:uniform}
\begin{figure}[t]
    \centering
    \includegraphics[width=\linewidth]{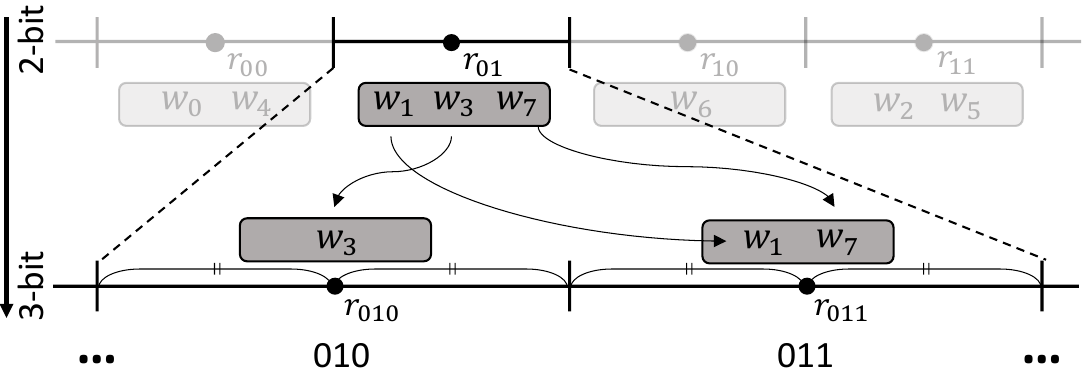}
    \caption{Incremental upscaling from 2-bit to 3-bit with uniform quantization methods.}
    \label{fig:upscale_uniform}
\end{figure}

In this section, we elaborate on our endeavor to apply incremental upscaling to the two state-of-the-art uniform quantization methods (GPTQ, AWQ), which ultimately results in failure. Figure~\ref{fig:upscale_uniform} illustrates the process of incremental upscaling with uniform quantization methods. Upscaling on uniform quantization methods requires that each bin in the quantization grid be divided equally; this is in contrast with non-uniform quantization, where each bin in the quantization grid can be divided into two new bins of arbitrary sizes. Naturally, each bin's midpoint becomes its representative value. With these premises, Appendix~\ref{appendix:gptq} and ~\ref{appendix:awq} discuss issues with upscaling GPTQ and AWQ, respectively.

\subsection{Incremental Upscaling with GPTQ}
\label{appendix:gptq}

\begin{algorithm}
\caption{Incremental Upscaling of GPTQ}
\label{algo:gptq}
\textbf{Input:} $\textbf{W}$, $\textbf{Q}_{n}$ \hfill // original weight, $n$-bit quantized weight

\textbf{Output:} $\textbf{Q}_{n+1}$ \hfill // ($n$+1)-bit quantized weight

\medskip

$\textbf{Q}_{n+1} = \textbf{0}_{K\times N}$

$\textbf{W}_{n+1} = \textbf{W}$

\For{$i = 0, ..., N$} {
    $\textbf{Q}_{n+1}[:,i] \leftarrow RTN_{n+1}(\textbf{W}_{n+1}[:,i])$
    
    \textcolor{red}{$\textbf{Q}_{n+1}[:,i]\leftarrow clamp(\textbf{Q}_{n+1}[:,i], \textbf{Q}_{n}[:,i]\times2, \textbf{Q}_{n}[:,i]\times2+1)$}

    $\textbf{E}\leftarrow compute\_err(\textbf{Q}_{n+1}[:,i], \textbf{W}_{n+1}[:,i])$

    $\textbf{W}_{n+1}[:,i:]\leftarrow compensate\_err(\textbf{W}_{n+1}[:,i:],  \textbf{E})$
}

\Return $\textbf{Q}_{n+1}$
\end{algorithm}

Algorithm \ref{algo:gptq} presents a modified version of GPTQ that additionally includes a clamping operation to preserve the essential weight-inheriting characteristic of the upscaling process. While clamping is necessary, it detrimentally forces weights to suboptimal values --- hence it is desirable for clamping to occur sparingly. As the algorithm processes each column of the weight matrix, it updates the error-compensated weights $W_{n+1}$ and, in correlation, generates the quantized weights $Q_{n+1}$. Therefore, to limit clamping, $W_{n+1}$ must evolve similarly to $W_{n}$, the error-compensated weights of the previous $n$-bit quantization, ensuring that $Q_{n+1}$ remains closely related to $Q_{n}$. However, this is not the observed case. $W_{n}$ and $W_{n+1}$, despite both starting as the original weight $W$, soon begin to diverge. The rounding functions $RTN_{n}$ and $RTN_{n+1}$ produce subtly different outcomes, leading to varied compensation for $W_{n}$ and $W_{n+1}$. Initially minor, this divergence grows over time, eventually necessitating clamping. Clamping then triggers a positive feedback loop, exacerbating disturbances in the upscaling process and further increasing clamping occurrences.

\begin{figure}[t]
    \centering
    \includegraphics[width=\linewidth]{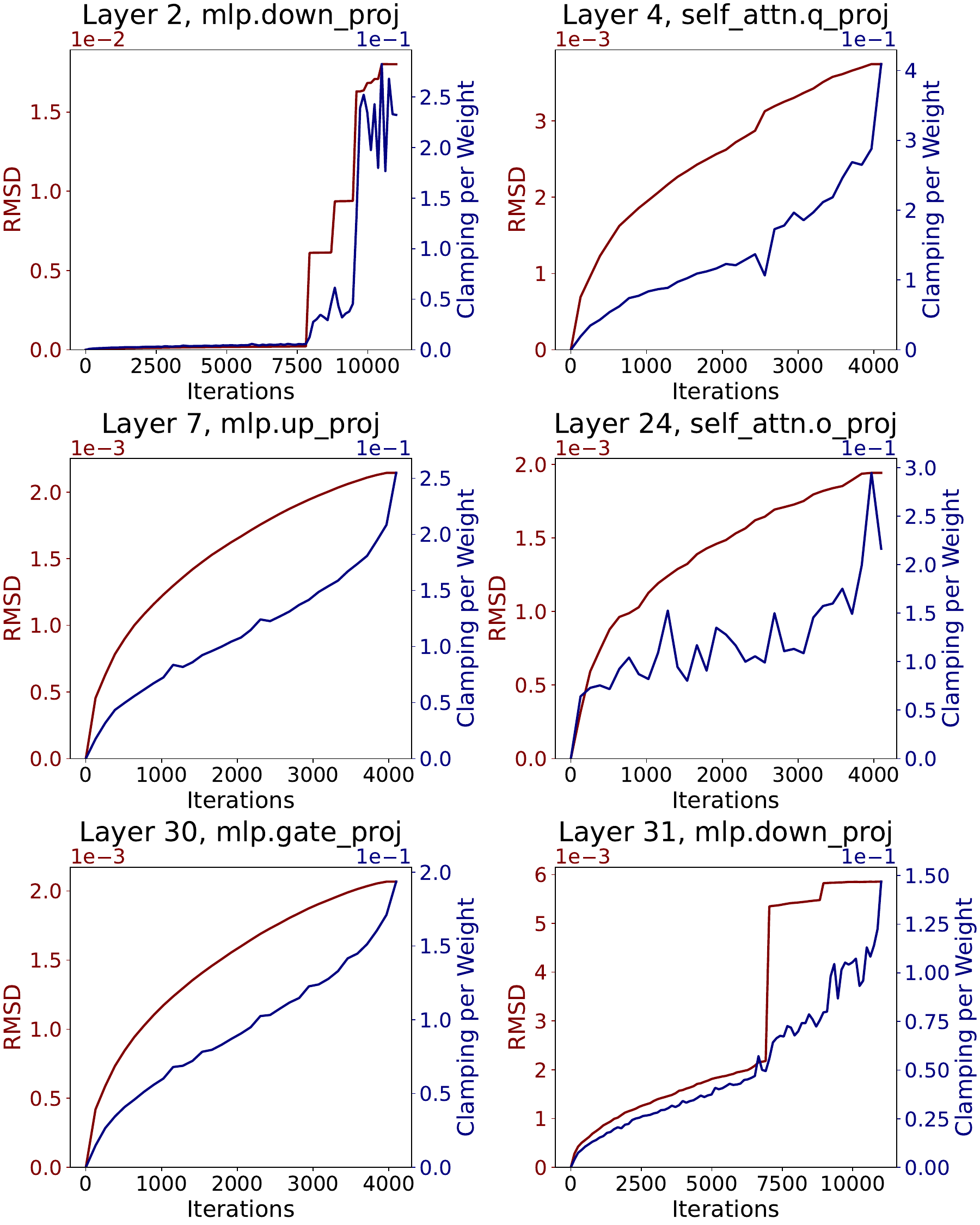}
    \caption{RMSD between the error-compensated weight matrices $W_3$ and $W_4$, along with the average clamping amount per weight, as the modified GPTQ algorithm progresses along the columns of Llama-2-7B.}
    \label{fig:gptq_analysis}
\end{figure}

In Figure~\ref{fig:gptq_analysis} we have randomly sampled six different weight matrices to demonstrate this runaway clamping effect. The error-compensated weight values $W_3$ and $W_4$ evolve differently in the seed 3-bit and upscaled 4-bit quantizations of the Llama-2-7B model. As the algorithm iterates over the columns, we plot the root-mean-square deviation (RMSD) of $W_4$ against $W_3$ on the left axis, along with the average clamping amount per weight value on the right axis. For all six matrices, the divergence between $W_{3}$ and $W_{4}$ starts small, but once the clamping amount starts to grow, it does so exponentially --- eventually leading to drastic errors.

The resulting instability of the upscaling process leads to the large quality drops shown in Table~\ref{table:appendix-uniform}, making GPTQ unfavorable for any-precison support. GPTQ-R refers to GPTQ with activation reordering.

\begin{figure}[t]
    \centering
    \includegraphics[width=\linewidth]{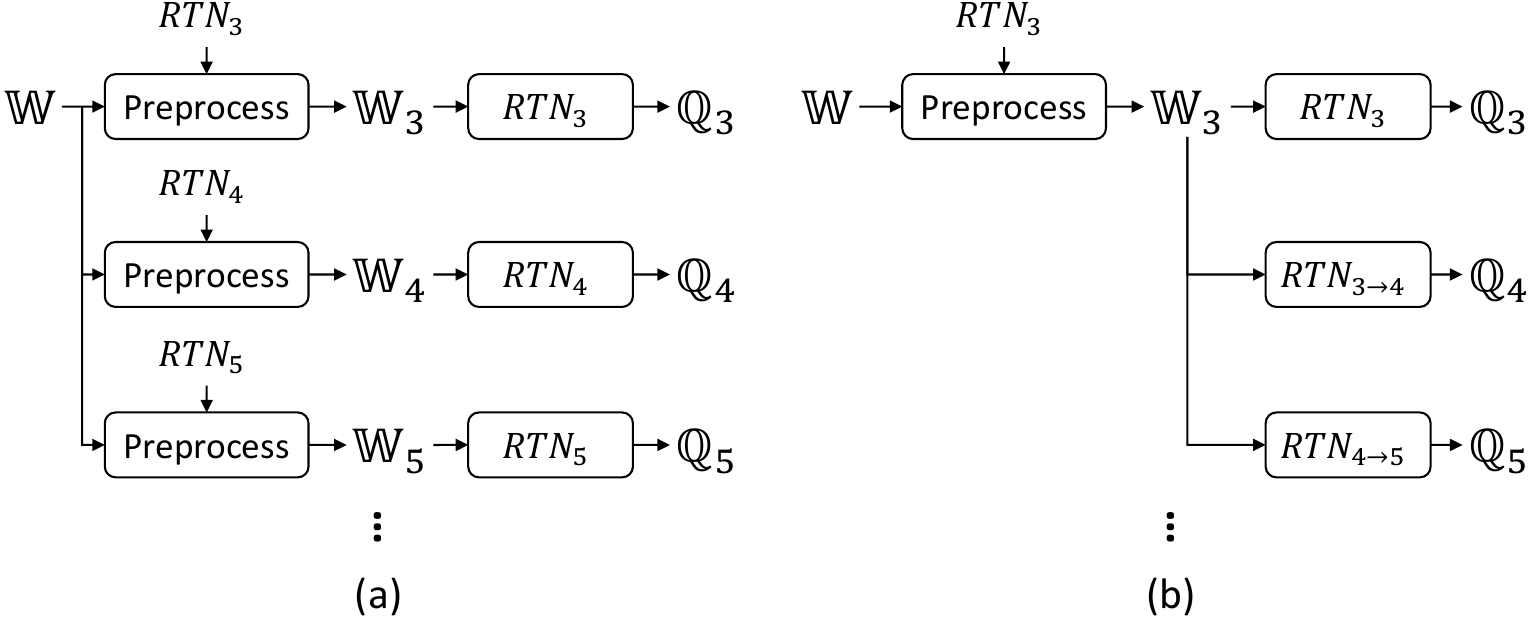}
    \caption{High-level abstraction of (a) original AWQ method and (b) AWQ with incremental upscaling.}
    \label{fig:awq}
\end{figure}

\subsection{Incremental Upscaling with AWQ}
\label{appendix:awq}
The AWQ process comprises two main steps, as depicted in Figure~\ref{fig:awq}(a). Initially, weight parameters, denoted as $\mathbb{W}$, undergo a preprocessing phase to become $\mathbb{W_n}$, involving channel-wise scaling and clipping adjustments tailored to the rounding function in use. For instance, in 3-bit quantization, the preprocessing involves $\mathbb{W}$ and a specific 3-bit rounding function, $RTN_{3}$, which segments the value range into $2^3$ equal parts for rounding. This step includes a grid search to determine the optimal scaling and clipping for minimal error post-rounding. Once preprocessing is finalized, applying $RTN_{3}$ to $\mathbb{W_n}$ yields the quantized weights, $\mathbb{Q_{3}}$. This methodology is adaptable for other bit-widths by substituting the appropriate rounding function, like $RTN_{4}$ or $RTN_{5}$, to suit the desired quantization precision.

When applying upscaling to AWQ, however, the preprocessing step used for the initial 3-bit model is reused across all bit-widths during upscaling, rather than tailoring preprocessing to each specific bit-width. This approach is depicted in Figure~\ref{fig:awq}(b) and is necessary to support any-precision quantization. For instance, upscaling from a 3-bit to a 4-bit model involves using the preprocessing result $\mathbb{W_{3}}$ from the 3-bit model and applying a function, $RTN_{3\rightarrow4}$, that redistributes values into $2^4$ bins by further dividing each 3-bit bin. This uniform preprocessing strategy is maintained for all upscaling steps to higher bit-widths. Although this method ensures consistency across quantizations, it relies on preprocessing factors optimized only for the seed model, which may not be ideal for upscaled models. The rounding functions used during upscaling, such as $RTN_{3\rightarrow4}$ and $RTN_{4\rightarrow5}$, differ from the seed model's $RTN_{3}$ in both the number of bins, the distribution of values, as well as whether a zero-point exists, which can lead to performance degradation as demonstrated in Table~\ref{table:appendix-uniform}.



\begin{sidewaystable*}[t]
    \centering
    \caption{Perplexity measured on Wikitext2, C4, and Penn Treebank for vanilla GPTQ, GPTQ-R and AWQ, as well as for their counterparts integrated with incremental scaling—namely, GPTQ+IU, GPTQ-R+IU, and AWQ+IU, respectively. 3-bit models are used as the seed models.}
    \label{table:appendix-uniform}
    \resizebox{\textwidth}{!} {
    \begin{tabular}{cc|ccc|ccc|ccc|ccc|ccc|ccc|}
         & ~ & \multicolumn{3}{c}{3-bit} & \multicolumn{3}{c}{4-bit} & \multicolumn{3}{c}{5-bit} & \multicolumn{3}{c}{6-bit} & \multicolumn{3}{c}{7-bit} & \multicolumn{3}{c|}{8-bit} \\ 
        ~ & ~ & Wiki & PTB & C4 & Wiki & PTB & C4 & Wiki & PTB & C4 & Wiki & PTB & C4 & Wiki & PTB & C4 & Wiki & PTB & C4 \\ \hline
        \multirow{6}{*}{\textbf{Llama-2-7B}} & GPTQ & 6.67 & 9.74 & 8.88 & 5.71 & 8.56 & 7.55 & 5.52 & 8.18 & 7.32 & 5.49 & 8.11 & 7.28 & 5.47 & 8.10 & 7.27 & 5.47 & 8.10 & 7.26 \\ 
        ~ & GPTQ+IU & - & - & - & 36.82 & 30.04 & 63.19 & 6.03 & 9.27 & 8.36 & 5.93 & 9.33 & 8.11 & 5.81 & 8.96 & 7.83 & 5.74 & 8.73 & 7.66 \\ \cline{2-20}
        ~ & GPTQ-R & 6.73 & 11.36 & 8.92 & 5.68 & 8.39 & 7.52 & 5.52 & 8.16 & 7.32 & 5.48 & 8.11 & 7.28 & 5.47 & 8.10 & 7.27 & 5.47 & 8.10 & 7.26 \\ 
        ~ & GPTQ-R+IU & - & - & - & 12.70 & 22.54 & 20.81 & 6.00 & 9.14 & 7.99 & 5.79 & 8.68 & 7.67 & 5.74 & 8.61 & 7.60 & 5.73 & 8.59 & 7.59 \\ \cline{2-20}
        ~ & AWQ & 6.24 & 9.03 & 8.30 & 5.59 & 8.27 & 7.44 & 5.50 & 8.14 & 7.31 & 5.48 & 8.11 & 7.28 & 5.47 & 8.09 & 7.27 & 5.47 & 8.10 & 7.26 \\ 
        ~ & AWQ+IU & - & - & - & INF & INF & INF & 22.50 & 39.37 & INF & 8.16 & 12.22 & 11.42 & 6.07 & 9.01 & 8.07 & 5.67 & 8.34 & 7.51 \\ \hline
        \multirow{6}{*}{\textbf{Mistral-7B}} & GPTQ & 8.93 & 12.02 & 13.19 & 5.43 & 8.78 & 8.64 & 5.29 & 8.52 & 8.45 & 5.26 & 8.49 & 8.40 & 5.25 & 8.46 & 8.38 & 5.25 & 8.46 & 8.38 \\ 
        ~ & GPTQ+IU & - & - & - & 16.73 & 32.65 & 20.67 & 5.93 & 10.58 & 9.10 & 5.47 & 8.86 & 8.61 & 5.42 & 8.71 & 8.54 & 5.40 & 8.68 & 8.52 \\ \cline{2-20}
        ~ & GPTQ-R & 6.47 & 10.85 & 9.74 & 5.40 & 8.67 & 8.60 & 5.27 & 8.52 & 8.43 & 5.25 & 8.47 & 8.39 & 5.25 & 8.46 & 8.38 & 5.25 & 8.46 & 8.38 \\ 
        ~ & GPTQ-R+IU & - & - & - & 15.69 & 30.34 & 26.39 & 5.71 & 9.17 & 8.89 & 5.40 & 8.77 & 8.61 & 5.36 & 8.69 & 8.56 & 5.35 & 8.67 & 8.54 \\ \cline{2-20}
        ~ & AWQ & 5.90 & 9.55 & 9.28 & 5.35 & 8.68 & 8.55 & 5.27 & 8.51 & 8.42 & 5.25 & 8.47 & 8.39 & 5.25 & 8.46 & 8.38 & 5.25 & 8.46 & 8.38 \\ 
        ~ & AWQ+IU & - & - & - & 20.07 & 45.75 & 32.88 & 7.68 & 15.33 & 12.78 & 5.83 & 9.68 & 9.24 & 5.42 & 8.79 & 8.62 & 5.34 & 8.62 & 8.52 \\ \hline
        \multirow{6}{*}{\textbf{OPT-6.7B}} & GPTQ & 11.58 & 13.43 & 13.94 & 10.90 & 12.64 & 13.25 & 10.80 & 12.59 & 13.09 & 10.77 & 12.55 & 13.07 & 10.81 & 12.52 & 13.05 & 10.84 & 12.52 & 13.05 \\ 
        ~ & GPTQ+IU & - & - & - & 12.40 & 14.13 & 14.44 & 11.18 & 12.77 & 13.33 & 10.95 & 12.62 & 13.15 & 10.86 & 12.62 & 13.12 & 10.84 & 12.59 & 13.09 \\ \cline{2-20}
        ~ & GPTQ-R & 11.70 & 13.75 & 13.94 & 11.03 & 12.79 & 13.20 & 10.86 & 12.59 & 13.07 & 10.84 & 12.57 & 13.05 & 10.84 & 12.55 & 13.05 & 10.84 & 12.52 & 13.05 \\ 
        ~ & GPTQ-R+IU & - & - & - & 12.52 & 13.99 & 14.38 & 11.23 & 12.95 & 13.35 & 11.03 & 12.87 & 13.23 & 10.96 & 12.84 & 13.20 & 10.96 & 12.84 & 13.20 \\ \cline{2-20}
        ~ & AWQ & 11.38 & 13.30 & 13.73 & 10.96 & 12.72 & 13.17 & 10.86 & 12.57 & 13.07 & 10.86 & 12.55 & 13.07 & 10.86 & 12.55 & 13.05 & 10.86 & 12.55 & 13.05 \\ 
        ~ & AWQ+IU & - & - & - & INF & INF & INF & INF & INF & INF & INF & INF & INF & INF & 55360.00 & INF & 27632.00 & 17296.00 & 28512.00 \\ \hline
        \multirow{6}{*}{\textbf{OPT-2.7B}} & GPTQ & 14.24 & 16.88 & 16.55 & 12.62 & 15.07 & 15.10 & 12.30 & 14.58 & 14.81 & 12.26 & 14.49 & 14.75 & 12.26 & 14.44 & 14.75 & 12.23 & 14.44 & 14.73 \\ 
        ~ & GPTQ+IU & - & - & - & 14.08 & 17.59 & 16.98 & 12.70 & 15.05 & 15.16 & 12.42 & 14.73 & 14.90 & 12.33 & 14.58 & 14.84 & 12.33 & 14.55 & 14.81 \\ \cline{2-20}
        ~ & GPTQ-R & 14.52 & 17.34 & 16.36 & 12.47 & 14.87 & 14.98 & 12.26 & 14.49 & 14.81 & 12.20 & 14.44 & 14.75 & 12.23 & 14.44 & 14.75 & 12.23 & 14.44 & 14.75 \\ 
        ~ & GPTQ-R+IU & - & - & - & 14.35 & 16.33 & 16.42 & 12.55 & 14.90 & 15.10 & 12.35 & 14.64 & 14.90 & 12.35 & 14.58 & 14.87 & 12.38 & 14.55 & 14.84 \\ \cline{2-20}
        ~ & AWQ & 13.56 & 16.14 & 15.95 & 12.70 & 14.66 & 14.98 & 12.52 & 14.47 & 14.81 & 12.49 & 14.44 & 14.75 & 12.49 & 14.44 & 14.75 & 12.47 & 14.44 & 14.75 \\ 
        ~ & AWQ+IU & - & - & - & INF & INF & INF & INF & 54080.00 & INF & 48864.00 & 50432.00 & 41152.00 & 37184.00 & 38976.00 & 31792.00 & 18400.00 & 16112.00 & 18256.00 \\ \hline
        \multirow{6}{*}{\textbf{OPT-1.3B}} & GPTQ & 17.48 & 19.66 & 19.17 & 15.05 & 16.88 & 17.19 & 14.93 & 16.59 & 17.25 & 14.70 & 16.30 & 16.81 & 14.75 & 16.33 & 16.91 & 14.81 & 16.36 & 16.95 \\ 
        ~ & GPTQ+IU & - & - & - & 16.98 & 18.88 & 19.05 & 15.37 & 16.98 & 17.25 & 14.98 & 16.66 & 16.91 & 14.90 & 16.55 & 16.81 & 14.90 & 16.55 & 16.78 \\ \cline{2-20}
        ~ & GPTQ-R & 16.98 & 18.91 & 18.61 & 14.95 & 16.63 & 16.91 & 14.70 & 16.36 & 16.66 & 14.70 & 16.27 & 16.59 & 14.66 & 16.23 & 16.59 & 14.64 & 16.23 & 16.59 \\ 
        ~ & GPTQ-R+IU & - & - & - & 16.39 & 18.00 & 18.28 & 15.22 & 16.75 & 17.08 & 15.02 & 16.66 & 16.91 & 14.93 & 16.52 & 16.88 & 14.90 & 16.52 & 16.88 \\ \cline{2-20}
        ~ & AWQ & 16.33 & 18.58 & 18.33 & 14.95 & 16.69 & 16.91 & 14.64 & 16.30 & 16.63 & 14.64 & 16.23 & 16.59 & 14.64 & 16.20 & 16.55 & 14.61 & 16.20 & 16.55 \\ 
        ~ & AWQ+IU & - & - & - & 7672.00 & 4074.00 & INF & 5192.00 & 3174.00 & INF & 3540.00 & 1970.00 & 2652.00 & 1559.00 & 780.50 & 1243.00 & 424.50 & 149.63 & 361.75 \\ \hline
    \end{tabular}
    }
\end{sidewaystable*}

\begin{sidewaystable*}[t]
    \centering
    \caption{Zero-shot accuracies on five tasks --- ARC-easy (ARC-e), ARC-challenge (ARC-c), HellaSwag, PIQA, and WinoGrande --- for vanilla SqueezeLLM (SqLLM) and SqueezeLLM integrated with incremental upscaling (SqLLM+IU) using a 3-bit seed model. We also report the difference between the two methods, and calculate the average accuracy across the five tasks.}
    \label{table:zero-shot}
    \resizebox{\linewidth}{!}{%
            \begin{tabular}{cc|cccccc|cccccc|cccccc|c|cccccc|}
                 & ~ & ARC-e & ARC-c & HellaSwag & PIQA & WinoGrande & Average & ARC-e & ARC-c & HellaSwag & PIQA & WinoGrande & Average & ARC-e & ARC-c & HellaSwag & PIQA & WinoGrande & Average & ~ & ARC-e & ARC-c & HellaSwag & PIQA & WinoGrande & Average \\ 
                \cline{1-20} \cline{22-27} \multirow{8}{*}{\textbf{Llama-2-7B}} & ~ & \multicolumn{6}{c|}{3-bit} & \multicolumn{6}{c|}{4-bit} & \multicolumn{6}{c|}{5-bit} & ~ & \multicolumn{6}{c|}{FP16}\\ 
                ~ & SqLLM & 70.3 \% & 41.8 \% & 73.0 \% & 77.9 \% & 67.8 \% & 66.2 \% & 73.8 \% & 45.5 \% & 75.3 \% & 78.8 \% & 68.0 \% & 68.3 \% & 74.3 \% & 45.4 \% & 75.8 \% & 78.8 \% & 68.8 \% & 68.6 \% & ~ & \multirow{7}{*}{74.6 \%} & \multirow{7}{*}{46.2 \%} & \multirow{7}{*}{76.0 \%} & \multirow{7}{*}{79.1 \%} & \multirow{7}{*}{69.0 \%} & \multirow{7}{*}{69.0 \%} \\
                ~ & SqLLM+IU & - & - & - & - & - & - & 73.5 \% & 44.2 \% & 75.3 \% & 78.7 \% & 69.1 \% & 68.2 \% & 74.5 \% & 45.8 \% & 75.8 \% & 78.9 \% & 68.9 \% & 68.8 \% & ~ & ~ & ~ & ~ & ~ & ~ & ~ \\ 
                ~ & $\Delta$ & - & - & - & - & - & - & -0.3 pp & -1.3 pp & +0.0 pp & -0.0 pp & +1.1 pp & -0.1 pp & +0.1 pp & +0.4 pp & +0.1 pp & +0.1 pp & +0.1 pp & +0.2 pp & ~ & ~ & ~ & ~ & ~ & ~ & ~ \\  \cline{2-20}
                ~ & ~ & \multicolumn{6}{c|}{6-bit} & \multicolumn{6}{c|}{7-bit} & \multicolumn{6}{c|}{8-bit} & ~ & ~ & ~ & ~ & ~ & ~ & ~ \\ 
                ~ & SqLLM & 74.5 \% & 46.2 \% & 75.9 \% & 78.8 \% & 68.8 \% & 68.8 \% & 74.6 \% & 45.9 \% & 76.0 \% & 79.2 \% & 69.1 \% & 68.9 \% & 74.4 \% & 46.1 \% & 76.0 \% & 79.1 \% & 69.0 \% & 68.9 \% & ~ & ~ & ~ & ~ & ~ & ~ & ~ \\ 
                ~ & SqLLM+IU & 74.6 \% & 46.2 \% & 76.1 \% & 79.1 \% & 68.6 \% & 68.9 \% & 74.4 \% & 46.1 \% & 76.0 \% & 79.1 \% & 69.1 \% & 68.9 \% & 74.3 \% & 46.3 \% & 76.0 \% & 79.1 \% & 69.1 \% & 69.0 \% & ~ & ~ & ~ & ~ & ~ & ~ & ~ \\ 
                ~ & $\Delta$ & +0.1 pp & 0.0 pp & +0.2 pp & +0.2 pp & -0.2 pp & +0.1 pp & -0.2 pp & +0.2 pp & -0.0 pp & -0.0 pp & +0.1 pp & 0.0 pp & -0.0 pp & +0.3 pp & -0.0 pp & 0.0 pp & +0.1 pp & +0.1 pp & ~ & ~ & ~ & ~ & ~ & ~ & ~ \\ 
                \cline{1-20} \cline{22-27} \multirow{8}{*}{\textbf{Mistral-7B}} & ~ & \multicolumn{6}{c|}{3-bit} & \multicolumn{6}{c|}{4-bit} & \multicolumn{6}{c|}{5-bit} & ~ & \multicolumn{6}{c|}{FP16}\\
                ~ & SqLLM & 76.0 \% & 49.4 \% & 78.1 \% & 81.6 \% & 72.5 \% & 71.5 \% & 77.9 \% & 52.1 \% & 80.3 \% & 81.6 \% & 73.9 \% & 73.2 \% & 79.1 \% & 53.7 \% & 80.9 \% & 81.8 \% & 73.7 \% & 73.8 \% & ~ & \multirow{7}{*}{79.6 \%} & \multirow{7}{*}{53.9 \%} & \multirow{7}{*}{81.0 \%} & \multirow{7}{*}{82.1 \%} & \multirow{7}{*}{73.9 \%} & \multirow{7}{*}{74.1 \%} \\ 
                ~ & SqLLM+IU & - & - & - & - & - & - & 78.2 \% & 52.3 \% & 80.3 \% & 81.9 \% & 74.0 \% & 73.3 \% & 79.2 \% & 53.5 \% & 80.8 \% & 82.2 \% & 73.6 \% & 73.8 \% & ~ & ~ & ~ & ~ & ~ & ~ & ~ \\ 
                ~ & $\Delta$ & - & - & - & - & - & - & +0.3 pp & +0.2 pp & -0.0 pp & +0.4 pp & +0.1 pp & +0.1 pp & +0.1 pp & -0.2 pp & -0.1 pp & +0.3 pp & -0.2 pp & 0.0 pp & ~ & ~ & ~ & ~ & ~ & ~ & ~ \\ \cline{2-20}
                ~ & ~ & \multicolumn{6}{c|}{6-bit} & \multicolumn{6}{c|}{7-bit} & \multicolumn{6}{c|}{8-bit} & ~ & ~ & ~ & ~ & ~ & ~ & ~ \\
                ~ & SqLLM & 79.5 \% & 53.4 \% & 81.0 \% & 82.1 \% & 74.1 \% & 74.0 \% & 79.5 \% & 53.8 \% & 81.0 \% & 82.2 \% & 73.8 \% & 74.1 \% & 79.6 \% & 54.1 \% & 81.1 \% & 82.2 \% & 73.9 \% & 74.2 \% & ~ & ~ & ~ & ~ & ~ & ~ & ~ \\ 
                ~ & SqLLM+IU & 79.4 \% & 53.7 \% & 81.0 \% & 82.0 \% & 73.6 \% & 74.0 \% & 79.4 \% & 54.2 \% & 81.1 \% & 82.3 \% & 73.9 \% & 74.2 \% & 79.3 \% & 53.9 \% & 81.1 \% & 82.1 \% & 74.0 \% & 74.1 \% & ~ & ~ & ~ & ~ & ~ & ~ & ~ \\ 
                ~ & $\Delta$ & -0.0 pp & +0.3 pp & 0.0 pp & -0.1 pp & -0.5 pp & 0.0 pp & -0.1 pp & +0.3 pp & +0.0 pp & +0.1 pp & +0.1 pp & +0.1 pp & -0.3 pp & -0.2 pp & -0.0 pp & -0.1 pp & +0.1 pp & -0.1 pp & ~ & ~ & ~ & ~ & ~ & ~ & ~ \\ 
                \cline{1-20} \cline{22-27} \multirow{8}{*}{\textbf{OPT-6.7B}} & ~ & \multicolumn{6}{c|}{3-bit} & \multicolumn{6}{c|}{4-bit} & \multicolumn{6}{c|}{5-bit} & ~ & \multicolumn{6}{c|}{FP16}\\ 
                ~ & SqLLM & 58.8 \% & 31.9 \% & 63.8 \% & 75.4 \% & 63.2 \% & 58.6 \% & 59.9 \% & 33.7 \% & 66.2 \% & 76.1 \% & 63.9 \% & 59.9 \% & 60.4 \% & 34.3 \% & 66.9 \% & 76.4 \% & 65.1 \% & 60.6 \% & ~ & \multirow{7}{*}{60.1 \%} & \multirow{7}{*}{34.6 \%} & \multirow{7}{*}{67.2 \%} & \multirow{7}{*}{76.7 \%} & \multirow{7}{*}{65.3 \%} & \multirow{7}{*}{60.8 \%} \\ 
                ~ & SqLLM+IU & - & - & - & - & - & - & 60.1 \% & 33.2 \% & 66.0 \% & 76.2 \% & 66.1 \% & 60.3 \% & 60.2 \% & 34.6 \% & 67.0 \% & 76.5 \% & 64.7 \% & 60.6 \% & ~ & ~ & ~ & ~ & ~ & ~ & ~ \\ 
                ~ & $\Delta$ & - & - & - & - & - & - & +0.2 pp & -0.5 pp & -0.2 pp & +0.1 pp & +2.3 pp & +0.4 pp & -0.2 pp & +0.3 pp & +0.1 pp & +0.1 pp & -0.4 pp & 0.0 pp & ~ & ~ & ~ & ~ & ~ & ~ & ~  \\  \cline{2-20}
                ~ & ~ & \multicolumn{6}{c|}{6-bit} & \multicolumn{6}{c|}{7-bit} & \multicolumn{6}{c|}{8-bit} & ~ & ~ & ~ & ~ & ~ & ~ & ~ \\ 
                ~ & SqLLM & 59.9 \% & 34.7 \% & 67.2 \% & 76.7 \% & 65.3 \% & 60.7 \% & 60.1 \% & 34.6 \% & 67.2 \% & 76.4 \% & 65.5 \% & 60.7 \% & 60.0 \% & 34.7 \% & 67.3 \% & 76.6 \% & 65.7 \% & 60.8 \% & ~ & ~ & ~ & ~ & ~ & ~ & ~ \\ 
                ~ & SqLLM+IU & 60.3 \% & 34.9 \% & 67.3 \% & 76.4 \% & 64.6 \% & 60.7 \% & 60.1 \% & 34.6 \% & 67.2 \% & 76.7 \% & 65.5 \% & 60.8 \% & 60.1 \% & 34.6 \% & 67.2 \% & 76.6 \% & 65.0 \% & 60.7 \% & ~ & ~ & ~ & ~ & ~ & ~ & ~ \\ 
                ~ & $\Delta$ & +0.4 pp & +0.2 pp & +0.2 pp & -0.3 pp & -0.6 pp & 0.0 pp & -0.0 pp & 0.0 pp & +0.0 pp & +0.3 pp & 0.0 pp & +0.1 pp & +0.2 pp & -0.1 pp & -0.0 pp & 0.0 pp & -0.7 pp & -0.1 pp & ~ & ~ & ~ & ~ & ~ & ~ & ~ \\ 
                \cline{1-20} \cline{22-27} \multirow{8}{*}{\textbf{OPT-2.7B}} & ~ & \multicolumn{6}{c|}{3-bit} & \multicolumn{6}{c|}{4-bit} & \multicolumn{6}{c|}{5-bit} & ~ & \multicolumn{6}{c|}{FP16}\\ 
                ~ & SqLLM & 53.2 \% & 30.5 \% & 56.0 \% & 72.6 \% & 59.4 \% & 54.3 \% & 53.5 \% & 31.3 \% & 59.2 \% & 74.2 \% & 60.6 \% & 55.8 \% & 53.9 \% & 31.2 \% & 60.1 \% & 74.8 \% & 61.3 \% & 56.3 \% & ~ & \multirow{7}{*}{54.4 \%} & \multirow{7}{*}{31.2 \%} & \multirow{7}{*}{60.7 \%} & \multirow{7}{*}{74.8 \%} & \multirow{7}{*}{60.8 \%} & \multirow{7}{*}{56.4 \%} \\ 
                ~ & SqLLM+IU & - & - & - & - & - & - & 54.4 \% & 31.1 \% & 59.0 \% & 74.1 \% & 60.1 \% & 55.7 \% & 54.3 \% & 30.7 \% & 60.0 \% & 74.4 \% & 60.9 \% & 56.1 \% & ~ & ~ & ~ & ~ & ~ & ~ & ~ \\ 
                ~ & $\Delta$ & - & - & - & - & - & - & +0.9 pp & -0.3 pp & -0.3 pp & -0.1 pp & -0.6 pp & -0.1 pp & +0.4 pp & -0.5 pp & -0.0 pp & -0.4 pp & -0.5 pp & -0.2 pp & ~ & ~ & ~ & ~ & ~ & ~ & ~ \\  \cline{2-20}
                ~ & ~ & \multicolumn{6}{c|}{6-bit} & \multicolumn{6}{c|}{7-bit} & \multicolumn{6}{c|}{8-bit} & ~ & ~ & ~ & ~ & ~ & ~ & ~ \\ 
                ~ & SqLLM & 54.1 \% & 31.2 \% & 60.4 \% & 74.7 \% & 60.9 \% & 56.2 \% & 54.4 \% & 30.9 \% & 60.5 \% & 74.7 \% & 61.2 \% & 56.3 \% & 54.2 \% & 31.1 \% & 60.6 \% & 74.8 \% & 61.0 \% & 56.3 \% & ~ & ~ & ~ & ~ & ~ & ~ & ~ \\ 
                ~ & SqLLM+IU & 54.3 \% & 31.0 \% & 60.4 \% & 74.8 \% & 60.5 \% & 56.2 \% & 54.3 \% & 31.5 \% & 60.5 \% & 74.6 \% & 61.2 \% & 56.4 \% & 54.3 \% & 31.3 \% & 60.6 \% & 74.8 \% & 60.8 \% & 56.3 \% & ~ & ~ & ~ & ~ & ~ & ~ & ~ \\ 
                ~ & $\Delta$ & +0.1 pp & -0.3 pp & -0.0 pp & +0.1 pp & -0.3 pp & 0.0 pp & -0.1 pp & +0.6 pp & +0.0 pp & -0.1 pp & 0.0 pp & +0.1 pp & +0.1 pp & +0.3 pp & -0.0 pp & 0.0 pp & -0.2 pp & 0.0 pp & ~ & ~ & ~ & ~ & ~ & ~ & ~ \\
                \cline{1-20} \cline{22-27} \multirow{8}{*}{\textbf{OPT-1.3B}} & ~ & \multicolumn{6}{c|}{3-bit} & \multicolumn{6}{c|}{4-bit} & \multicolumn{6}{c|}{5-bit} & ~ & \multicolumn{6}{c|}{FP16} \\
                ~ & SqLLM & 48.3 \% & 27.5 \% & 49.6 \% & 71.8 \% & 56.3 \% & 50.7 \% & 51.5 \% & 28.5 \% & 52.9 \% & 71.2 \% & 59.3 \% & 52.7 \% & 51.2 \% & 29.0 \% & 53.6 \% & 71.9 \% & 59.0 \% & 52.9 \% & ~ & \multirow{7}{*}{51.0 \%} & \multirow{7}{*}{29.6 \%} & \multirow{7}{*}{53.7 \%} & \multirow{7}{*}{72.5 \%} & \multirow{7}{*}{60.0 \%} & \multirow{7}{*}{53.3 \%} \\ 
                ~ & SqLLM+IU & - & - & - & - & - & - & 49.8 \% & 29.9 \% & 53.0 \% & 71.7 \% & 58.6 \% & 52.6 \% & 50.7 \% & 29.6 \% & 53.5 \% & 72.4 \% & 59.1 \% & 53.1 \% & ~ & ~ & ~ & ~ & ~ & ~ & ~ \\ 
                ~ & $\Delta$ & - & - & - & - & - & - & -1.7 pp & +1.4 pp & +0.0 pp & +0.4 pp & -0.7 pp & -0.1 pp & -0.5 pp & +0.6 pp & -0.0 pp & +0.5 pp & +0.2 pp & +0.2 pp & ~ & ~ & ~ & ~ & ~ & ~ & ~ \\ \cline{2-20}
                ~ & ~ & \multicolumn{6}{c|}{6-bit} & \multicolumn{6}{c|}{7-bit} & \multicolumn{6}{c|}{8-bit} & ~ & ~ & ~ & ~ & ~ & ~ & ~ \\ 
                ~ & SqLLM & 51.4 \% & 30.0 \% & 53.7 \% & 72.4 \% & 59.4 \% & 53.4 \% & 50.9 \% & 29.7 \% & 53.7 \% & 72.4 \% & 60.1 \% & 53.4 \% & 51.0 \% & 29.5 \% & 53.7 \% & 72.4 \% & 59.9 \% & 53.3 \% & ~ & ~ & ~ & ~ & ~ & ~ & ~ \\ 
                ~ & SqLLM+IU & 51.0 \% & 29.7 \% & 53.7 \% & 72.5 \% & 59.4 \% & 53.3 \% & 50.9 \% & 29.8 \% & 53.7 \% & 72.6 \% & 60.2 \% & 53.4 \% & 51.1 \% & 29.8 \% & 53.7 \% & 72.4 \% & 59.4 \% & 53.3 \% & ~ & ~ & ~ & ~ & ~ & ~ & ~ \\ 
                ~ & $\Delta$ & -0.4 pp & -0.3 pp & 0.0 pp & +0.1 pp & 0.0 pp & -0.1 pp & -0.0 pp & +0.1 pp & -0.0 pp & +0.3 pp & +0.1 pp & 0.0 pp & +0.1 pp & +0.3 pp & -0.0 pp & 0.0 pp & -0.5 pp & 0.0 pp & ~ & ~ & ~ & ~ & ~ & ~ & ~ \\ \cline{1-20} \cline{22-27}
    \end{tabular}
    }
\end{sidewaystable*}

\begin{table*}[t]
    \centering
    \caption{Latency of our kernel on matrix-vector multiplication of various dimensions.}
    \label{table:various-sizes}
    \resizebox{\textwidth}{!}{%
    \begin{tabular}{c|c|c|cccccc|c}
    \hline
        GPU & Model & Matrix Size & 3-bit & 4-bit & 5-bit & 6-bit & 7-bit & 8-bit & FP16 \\ \hline
        \multirow{13}{*}{RTX 4090} & \multirow{3}{*}{OPT-1.3B} & 2048 x 2048 & 4.73 & 5.95 & 6.63 & 7.13 & 7.95 & 9.06 & 12.16 \\
         & & 8192 x 2048 & 9.72 & 12.65 & 14.82 & 17.82 & 22.42 & 27.92 & 38.66 \\
         & & 2048 x 8192 & 9.78 & 12.88 & 15.60 & 18.15 & 21.45 & 24.14 & 42.62 \\ \cline{2-10}
         & \multirow{3}{*}{OPT-2.7B} & 2560 x 2560 & 5.89 & 7.45 & 8.26 & 9.00 & 10.16 & 11.77 & 19.55 \\
         & & 10240 x 2560 & 14.29 & 20.34 & 24.30 & 28.26 & 32.81 & 38.80 & 58.50 \\
         & & 2560 x 10240 & 13.88 & 18.83 & 22.68 & 26.80 & 37.60 & 40.59 & 60.90 \\ \cline{2-10}
         & \multirow{3}{*}{Llama-2-7B} & 4096 x 4096 & 9.67 & 12.73 & 14.80 & 17.72 & 20.64 & 24.69 & 38.59 \\
         & & 11008 x 4096 & 21.90 & 29.80 & 36.26 & 41.95 & 48.13 & 57.44 & 102.34 \\
         & & 4096 x 11008 & 23.16 & 30.09 & 37.56 & 45.19 & 49.55 & 56.98 & 103.04 \\ \cline{2-10}
         & \multirow{2}{*}{Mistral-7B} & 14336 x 4096 & 28.73 & 39.00 & 46.13 & 52.06 & 62.16 & 74.44 & 132.38 \\
         & & 4096 x 14336 & 27.69 & 36.43 & 43.65 & 50.57 & 61.83 & 69.77 & 134.78 \\ \cline{2-10}
         & \multirow{2}{*}{OPT-6.7B} & 16384 x 4096 & 30.72 & 42.12 & 49.84 & 59.13 & 69.99 & 86.55 & 150.34 \\
         & & 4096 x 16384 & 30.80 & 40.63 & 48.83 & 57.47 & 69.96 & 79.35 & 156.70 \\ \hline
        \multirow{13}{*}{RTX 4070 Laptop} & \multirow{3}{*}{OPT-1.3B} & 2048 x 2048 & 9.64 & 12.63 & 15.04 & 17.79 & 20.85 & 24.79 & 36.74 \\
         & & 8192 x 2048 & 28.93 & 38.26 & 47.46 & 58.08 & 71.17 & 87.97 & 142.05 \\
         & & 2048 x 8192 & 29.27 & 39.49 & 48.17 & 56.01 & 68.52 & 77.58 & 145.76 \\ \cline{2-10}
         & \multirow{3}{*}{OPT-2.7B} & 2560 x 2560 & 13.92 & 18.94 & 22.56 & 26.00 & 29.94 & 35.26 & 57.66 \\
         & & 10240 x 2560 & 43.90 & 60.75 & 75.12 & 90.90 & 107.90 & 134.34 & 219.30 \\
         & & 2560 x 10240 & 42.78 & 58.30 & 71.78 & 85.25 & 102.77 & 116.97 & 220.10 \\ \cline{2-10}
         & \multirow{3}{*}{Llama-2-7B} & 4096 x 4096 & 28.57 & 38.12 & 47.16 & 56.55 & 67.48 & 80.65 & 142.05 \\
         & & 11008 x 4096 & 71.71 & 96.22 & 120.09 & 149.99 & 179.50 & 214.07 & 369.12 \\
         & & 4096 x 11008 & 72.21 & 104.42 & 125.35 & 151.54 & 179.48 & 204.47 & 382.21 \\ \cline{2-10}
         & \multirow{2}{*}{Mistral-7B} & 14336 x 4096 & 92.42 & 124.22 & 161.02 & 193.62 & 232.61 & 276.84 & 478.08 \\
         & & 4096 x 14336 & 92.53 & 134.28 & 164.77 & 197.07 & 223.79 & 258.43 & 484.10 \\ \cline{2-10}
         & \multirow{2}{*}{OPT-6.7B} & 16384 x 4096 & 105.06 & 145.05 & 182.29 & 220.29 & 261.84 & 311.81 & 545.12 \\
         & & 4096 x 16384 & 104.97 & 151.25 & 191.31 & 224.63 & 253.72 & 291.27 & 546.66 \\ \hline
        \multirow{13}{*}{Jetson AGX Orin} & \multirow{3}{*}{OPT-1.3B} & 2048 x 2048 & 20.81 & 25.88 & 28.88 & 31.58 & 34.70 & 39.86 & 57.66 \\
         & & 8192 x 2048 & 58.37 & 72.80 & 85.00 & 94.15 & 106.76 & 126.68 & 198.18 \\
         & & 2048 x 8192 & 53.22 & 67.72 & 81.98 & 88.53 & 108.69 & 112.26 & 207.14 \\ \cline{2-10}
         & \multirow{3}{*}{OPT-2.7B} & 2560 x 2560 & 30.49 & 38.88 & 43.60 & 47.57 & 51.97 & 58.90 & 84.83 \\
         & & 10240 x 2560 & 92.90 & 118.15 & 142.74 & 154.36 & 171.82 & 191.68 & 304.90 \\
         & & 2560 x 10240 & 77.66 & 99.02 & 119.97 & 130.06 & 148.89 & 170.22 & 322.56 \\ \cline{2-10}
         & \multirow{3}{*}{Llama-2-7B} & 4096 x 4096 & 53.87 & 68.54 & 80.96 & 88.65 & 98.43 & 116.01 & 206.88 \\
         & & 11008 x 4096 & 128.10 & 162.52 & 198.02 & 217.67 & 246.35 & 291.93 & 538.21 \\
         & & 4096 x 11008 & 127.33 & 186.92 & 218.29 & 219.58 & 256.50 & 297.60 & 553.73 \\ \cline{2-10}
         & \multirow{2}{*}{Mistral-7B} & 14336 x 4096 & 164.20 & 209.12 & 255.67 & 282.39 & 321.54 & 377.20 & 691.33 \\
         & & 4096 x 14336 & 158.01 & 206.10 & 252.93 & 284.34 & 328.82 & 359.19 & 717.70 \\ \cline{2-10}
         & \multirow{2}{*}{OPT-6.7B} & 16384 x 4096 & 186.25 & 237.60 & 289.68 & 321.12 & 361.60 & 430.34 & 788.32 \\
         & & 4096 x 16384 & 179.27 & 236.02 & 288.04 & 312.59 & 352.66 & 404.15 & 805.25 \\ \hline
    \end{tabular}
    }
\end{table*}

\begin{table*}[t]
    \centering
    \caption{Comparison of matrix-vector multiplication latency against existing uniform quantization kernels. The best results are highlighted for each case.}
    \label{table:compare-uniform}
    \resizebox{\textwidth}{!}{%
    \begin{tabular}{c|c|ccc|ccc|ccc}
    \hline
        \multicolumn{2}{c|}{GPU} & \multicolumn{3}{c|}{RTX 4090} & \multicolumn{3}{c|}{RTX 4070 Laptop} & \multicolumn{3}{c}{Jetson AGX Orin} \\ \hline
        \multicolumn{2}{c|}{Matrix Size} & 4096x4096 & 11008x4096 & 4096x11008 & 4096x4096 & 11008x4096 & 4096x11008 & 4096x4096 & 11008x4096 & 4096x11008 \\ \hline
         & ExLlamaV2 & 12.15 & 24.89 & 24.95 & 31.8 & 74.72 & 74.94 & 67.1 & 137.5 & 134.45 \\
         & LUT-GEMM & 12.52 & 25.03 & 25.39 & 35.04 & 83.23 & 85.05 & 61.9 & 131.66 & 140.36 \\
        \rowcolor{cyan!10}\multirow{-3}{*}{\cellcolor{white}3-bit} & Ours & \textbf{9.67} & \textbf{21.90} & \textbf{23.16} & \textbf{28.57} & \textbf{71.71} & \textbf{72.21} & \textbf{53.87} & \textbf{128.10} & \textbf{127.33} \\ \hline
         & ExLlamaV2 & 14.89 & 31.11 & 31.88 & 41.57 & 97.25 & \textbf{97.28} & 81.1 & 171.61 & 172.68 \\
         & AWQ & 13.89 & 30.96 & 30.21 & 41.04 & 100.83 & 105.26 & 73.12 & 176.69 & 177.77 \\
         & TRT-LLM & \textbf{12.58} & \textbf{29.34} & \textbf{27.94} & 39.58 & 96.86 & 98.98 & - & - & - \\
         & LUT-GEMM & 14.78 & 31.73 & 31.78 & 45.35 & 108.33 & 109.87 & 71.64 & 164.27 & \textbf{170.49} \\
        \rowcolor{cyan!10}\multirow{-5}{*}{\cellcolor{white}4-bit} & Ours & 12.73 & 29.8 & 30.09 & \textbf{38.12} & \textbf{96.22} & 104.42 & \textbf{68.54} & \textbf{162.52} & 186.92 \\ \hline
         & ExLlamaV2 & 16.1 & \textbf{35.92} & \textbf{36.11} & 48.39 & 120.49 & \textbf{120.75} & 90.68 & \textbf{194.54} & \textbf{194.19} \\
         & LUT-GEMM & 17.52 & 38.39 & 38.72 & 54.17 & 137.43 & 138.76 & 87.37 & 203.21 & 221.47 \\
        \rowcolor{cyan!10}\multirow{-3}{*}{\cellcolor{white}5-bit} & Ours & \textbf{14.80} & 36.26 & 37.56 & \textbf{47.16} & \textbf{120.09} & 125.35 & \textbf{80.96} & 198.02 & 218.29 \\ \hline
         & ExLlamaV2 & 19.74 & 43.06 & \textbf{43.48} & 58.15 & \textbf{147.21} & \textbf{147.84} & 101.75 & 228.82 & 223.38 \\
         & LUT-GEMM & 19.37 & 45.14 & 46.26 & 66.45 & 165.59 & 166.18 & 100.46 & 236.18 & 251.02 \\
        \rowcolor{cyan!10}\multirow{-3}{*}{\cellcolor{white}6-bit} & Ours & \textbf{17.72} & \textbf{41.95} & 45.19 & \textbf{56.55} & 149.99 & 151.54 & \textbf{88.65} & \textbf{217.67} & \textbf{219.58} \\ \hline
         & LUT-GEMM & 22.54 & 51.81 & 51.54 & 73.23 & 191.64 & 192.36 & 117.89 & 273.49 & 301.48 \\
        \rowcolor{cyan!10}\multirow{-2}{*}{\cellcolor{white}7-bit} & Ours & \textbf{20.64} & \textbf{48.13} & \textbf{49.55} & \textbf{67.48} & \textbf{179.50} & \textbf{179.48} & \textbf{98.43} & \textbf{246.35} & \textbf{256.50} \\ \hline
         & ExLlamaV2 & \textbf{23.52} & \textbf{53.07} & \textbf{53.55} & \textbf{75.79} & \textbf{193.00} & \textbf{192.91} & 129.96 & 303.93 & \textbf{292.41} \\
         & LUT-GEMM & 25.04 & 59.27 & 59.01 & 83.61 & 216.59 & 217.3 & 125.12 & 308.45 & 328.92 \\
        \rowcolor{cyan!10}\multirow{-3}{*}{\cellcolor{white}8-bit} & Ours & 24.69 & 57.44 & 56.98 & 80.65 & 214.07 & 204.47 & \textbf{116.01} & \textbf{291.93} & 297.6 \\ \hline
    \end{tabular}
    }
\end{table*}

\begin{table*}[t]
    \centering
    \caption{Matrix-matrix multiplication latency of our kernel with batch sizes of 2, 4 and 8. We also present the rate of latency increase against single-batch for each case. Cases with latency increases of less than 30\% are highlighted in blue, while those which exhibit more than a twofold increase are highlighted in red.}
    \label{table:multi-batch}
    \resizebox{\textwidth}{!}{%
    \begin{tabular}{c|c|ccc|ccc|ccc}
    \hline
        \multicolumn{2}{c|}{GPU} & \multicolumn{3}{c|}{RTX 4090} & \multicolumn{3}{c|}{RTX 4070 Laptop} & \multicolumn{3}{c}{Jetson AGX Orin} \\ \hline
        ~ & BS & 4096x4096 & 11008x4096 & 4096x11008 & 4096x4096 & 11008x4096 & 4096x11008 & 4096x4096 & 11008x4096 & 4096x11008 \\ \hline
        \multirow{8}{*}{3-bit} & \multirow{2}{*}{1} & 9.67 & 21.90 & 23.16 & 28.57 & 71.71 & 72.21 & 53.87 & 128.10 & 127.33 \\
        ~ & ~ & ($\times$1.00) & ($\times$1.00) & ($\times$1.00) & ($\times$1.00) & ($\times$1.00) & ($\times$1.00) & ($\times$1.00) & ($\times$1.00) & ($\times$1.00) \\ \cline{2-11}
        ~ & \multirow{2}{*}{2} & 10.73 & 23.77 & 28.36 & 33.72 & 72.87 & 88.15 & 59.81 & 141.49 & 152.23 \\
        ~ & ~ & {\color{blue} ($\times$1.11)} & {\color{blue} ($\times$1.09)} & {\color{blue} ($\times$1.22)} & {\color{blue} ($\times$1.18)} & {\color{blue} ($\times$1.02)} & {\color{blue} ($\times$1.22)} & {\color{blue} ($\times$1.11)} & {\color{blue} ($\times$1.10)} & {\color{blue} ($\times$1.20)} \\
        ~ & \multirow{2}{*}{4} & 11.98 & 32.27 & 30.45 & 32.65 & 76.79 & 80.50 & 87.93 & 215.84 & 216.02 \\
        ~ & ~ & {\color{blue} ($\times$1.24)} & ($\times$1.47) & ($\times$1.31) & {\color{blue} ($\times$1.14)} & {\color{blue} ($\times$1.07)} & {\color{blue} ($\times$1.11)} & ($\times$1.63) & ($\times$1.68) & ($\times$1.70) \\
        ~ & \multirow{2}{*}{8} & 15.65 & 44.29 & 40.55 & 51.12 & 121.58 & 125.21 & 151.02 & 388.55 & 378.78 \\
        ~ & ~ & ($\times$1.62) & {\color{red} ($\times$2.02)} & ($\times$1.75) & ($\times$1.79) & ($\times$1.70) & ($\times$1.73) & {\color{red} ($\times$2.80)} & {\color{red} ($\times$3.03)} & {\color{red} ($\times$2.97)} \\ \hline
        \multirow{8}{*}{4-bit} & \multirow{2}{*}{1} & 12.73 & 29.80 & 30.09 & 38.12 & 96.22 & 104.42 & 68.54 & 162.52 & 186.92 \\
        ~ & ~ & ($\times$1.00) & ($\times$1.00) & ($\times$1.00) & ($\times$1.00) & ($\times$1.00) & ($\times$1.00) & ($\times$1.00) & ($\times$1.00) & ($\times$1.00) \\ \cline{2-11}
        ~ & \multirow{2}{*}{2} & 13.93 & 37.92 & 39.12 & 41.38 & 100.14 & 107.57 & 75.91 & 183.80 & 191.78 \\
        ~ & ~ & {\color{blue} ($\times$1.09)} & {\color{blue} ($\times$1.27)} & {\color{blue} ($\times$1.30)} & {\color{blue} ($\times$1.09)} & {\color{blue} ($\times$1.04)} & {\color{blue} ($\times$1.03)} & {\color{blue} ($\times$1.11)} & {\color{blue} ($\times$1.13)} & {\color{blue} ($\times$1.03)} \\
        ~ & \multirow{2}{*}{4} & 14.93 & 38.53 & 39.17 & 41.99 & 101.35 & 106.85 & 95.10 & 232.20 & 240.12 \\
        ~ & ~ & {\color{blue} ($\times$1.17)} & {\color{blue} ($\times$1.29)} & {\color{blue} ($\times$1.30)} & {\color{blue} ($\times$1.10)} & {\color{blue} ($\times$1.05)} & {\color{blue} ($\times$1.02)} & ($\times$1.39) & ($\times$1.43) & {\color{blue} ($\times$1.28)} \\
        ~ & \multirow{2}{*}{8} & 17.53 & 48.71 & 45.45 & 57.37 & 135.49 & 140.48 & 175.37 & 452.44 & 430.61 \\
        ~ & ~ & ($\times$1.38) & ($\times$1.63) & ($\times$1.51) & ($\times$1.51) & ($\times$1.41) & ($\times$1.35) & {\color{red} ($\times$2.56)} & {\color{red} ($\times$2.78)} & {\color{red} ($\times$2.30)} \\ \hline
        \multirow{8}{*}{5-bit} & \multirow{2}{*}{1} & 14.80 & 36.26 & 37.56 & 47.16 & 120.09 & 125.35 & 80.96 & 198.02 & 218.29 \\
        ~ & ~ & ($\times$1.00) & ($\times$1.00) & ($\times$1.00) & ($\times$1.00) & ($\times$1.00) & ($\times$1.00) & ($\times$1.00) & ($\times$1.00) & ($\times$1.00) \\ \cline{2-11}
        ~ & \multirow{2}{*}{2} & 15.40 & 36.33 & 42.67 & 49.26 & 122.81 & 138.92 & 85.92 & 208.13 & 218.24 \\
        ~ & ~ & {\color{blue} ($\times$1.04)} & {\color{blue} ($\times$1.00)} & {\color{blue} ($\times$1.14)} & {\color{blue} ($\times$1.04)} & {\color{blue} ($\times$1.02)} & {\color{blue} ($\times$1.11)} & {\color{blue} ($\times$1.06)} & {\color{blue} ($\times$1.05)} & {\color{blue} ($\times$1.00)} \\
        ~ & \multirow{2}{*}{4} & 18.33 & 42.45 & 47.52 & 51.62 & 124.37 & 135.16 & 104.15 & 254.89 & 263.17 \\
        ~ & ~ & {\color{blue} ($\times$1.24)} & {\color{blue} ($\times$1.17)} & {\color{blue} ($\times$1.27)} & {\color{blue} ($\times$1.09)} & {\color{blue} ($\times$1.04)} & {\color{blue} ($\times$1.08)} & {\color{blue} ($\times$1.29)} & {\color{blue} ($\times$1.29)} & {\color{blue} ($\times$1.21)} \\
        ~ & \multirow{2}{*}{8} & 22.53 & 50.97 & 54.47 & 61.49 & 134.38 & 152.76 & 136.60 & 334.85 & 446.29 \\
        ~ & ~ & ($\times$1.52) & ($\times$1.41) & ($\times$1.45) & {\color{blue} ($\times$1.30)} & {\color{blue} ($\times$1.12)} & {\color{blue} ($\times$1.22)} & ($\times$1.69) & ($\times$1.69) & {\color{red} ($\times$2.04)} \\ \hline
        \multirow{8}{*}{6-bit} & \multirow{2}{*}{1} & 17.72 & 41.95 & 45.19 & 56.55 & 149.99 & 151.54 & 88.65 & 217.67 & 219.58 \\
        ~ & ~ & ($\times$1.00) & ($\times$1.00) & ($\times$1.00) & ($\times$1.00) & ($\times$1.00) & ($\times$1.00) & ($\times$1.00) & ($\times$1.00) & ($\times$1.00) \\ \cline{2-11}
        ~ & \multirow{2}{*}{2} & 17.89 & 43.41 & 48.21 & 61.90 & 158.64 & 173.25 & 93.35 & 223.61 & 236.13 \\
        ~ & ~ & {\color{blue} ($\times$1.01)} & {\color{blue} ($\times$1.03)} & {\color{blue} ($\times$1.07)} & {\color{blue} ($\times$1.09)} & {\color{blue} ($\times$1.06)} & {\color{blue} ($\times$1.14)} & {\color{blue} ($\times$1.05)} & {\color{blue} ($\times$1.03)} & {\color{blue} ($\times$1.08)} \\
        ~ & \multirow{2}{*}{4} & 20.61 & 47.45 & 53.43 & 59.48 & 153.59 & 161.73 & 110.99 & 271.59 & 284.77 \\
        ~ & ~ & {\color{blue} ($\times$1.16)} & {\color{blue} ($\times$1.13)} & {\color{blue} ($\times$1.18)} & {\color{blue} ($\times$1.05)} & {\color{blue} ($\times$1.02)} & {\color{blue} ($\times$1.07)} & {\color{blue} ($\times$1.25)} & {\color{blue} ($\times$1.25)} & {\color{blue} ($\times$1.30)} \\
        ~ & \multirow{2}{*}{8} & 25.19 & 55.84 & 60.69 & 70.47 & 165.17 & 177.89 & 148.84 & 362.45 & 502.36 \\
        ~ & ~ & ($\times$1.42) & ($\times$1.33) & ($\times$1.34) & {\color{blue} ($\times$1.25)} & {\color{blue} ($\times$1.10)} & {\color{blue} ($\times$1.17)} & ($\times$1.68) & ($\times$1.67) & {\color{red} ($\times$2.29)} \\ \hline
        \multirow{8}{*}{7-bit} & \multirow{2}{*}{1} & 20.64 & 48.13 & 49.55 & 67.48 & 179.50 & 179.48 & 98.43 & 246.35 & 256.50 \\
        ~ & ~ & ($\times$1.00) & ($\times$1.00) & ($\times$1.00) & ($\times$1.00) & ($\times$1.00) & ($\times$1.00) & ($\times$1.00) & ($\times$1.00) & ($\times$1.00) \\ \cline{2-11}
        ~ & \multirow{2}{*}{2} & 21.74 & 49.79 & 56.18 & 70.92 & 183.65 & 189.63 & 105.25 & 255.30 & 268.83 \\
        ~ & ~ & {\color{blue} ($\times$1.05)} & {\color{blue} ($\times$1.03)} & {\color{blue} ($\times$1.13)} & {\color{blue} ($\times$1.05)} & {\color{blue} ($\times$1.02)} & {\color{blue} ($\times$1.06)} & {\color{blue} ($\times$1.07)} & {\color{blue} ($\times$1.04)} & {\color{blue} ($\times$1.05)} \\
        ~ & \multirow{2}{*}{4} & 23.92 & 52.73 & 60.77 & 69.44 & 181.75 & 185.54 & 122.35 & 295.96 & 319.96 \\
        ~ & ~ & {\color{blue} ($\times$1.16)} & {\color{blue} ($\times$1.10)} & {\color{blue} ($\times$1.23)} & {\color{blue} ($\times$1.03)} & {\color{blue} ($\times$1.01)} & {\color{blue} ($\times$1.03)} & {\color{blue} ($\times$1.24)} & {\color{blue} ($\times$1.20)} & {\color{blue} ($\times$1.25)} \\
        ~ & \multirow{2}{*}{8} & 28.15 & 59.43 & 67.21 & 79.33 & 193.08 & 205.41 & 157.09 & 384.32 & 517.72 \\
        ~ & ~ & ($\times$1.36) & {\color{blue} ($\times$1.23)} & ($\times$1.36) & {\color{blue} ($\times$1.18)} & {\color{blue} ($\times$1.08)} & {\color{blue} ($\times$1.14)} & ($\times$1.60) & ($\times$1.56) & {\color{red} ($\times$2.02)} \\ \hline
        \multirow{8}{*}{8-bit} & \multirow{2}{*}{1} & 24.69 & 57.44 & 56.98 & 80.65 & 214.07 & 204.47 & 116.01 & 291.93 & 297.60 \\
        ~ & ~ & ($\times$1.00) & ($\times$1.00) & ($\times$1.00) & ($\times$1.00) & ($\times$1.00) & ($\times$1.00) & ($\times$1.00) & ($\times$1.00) & ($\times$1.00) \\ \cline{2-11}
        ~ & \multirow{2}{*}{2} & 26.64 & 58.35 & 67.43 & 81.81 & 220.69 & 218.08 & 123.00 & 300.90 & 319.13 \\
        ~ & ~ & {\color{blue} ($\times$1.08)} & {\color{blue} ($\times$1.02)} & {\color{blue} ($\times$1.18)} & {\color{blue} ($\times$1.01)} & {\color{blue} ($\times$1.03)} & {\color{blue} ($\times$1.07)} & {\color{blue} ($\times$1.06)} & {\color{blue} ($\times$1.03)} & {\color{blue} ($\times$1.07)} \\
        ~ & \multirow{2}{*}{4} & 27.46 & 59.28 & 68.35 & 82.97 & 217.70 & 212.40 & 136.70 & 335.56 & 351.07 \\
        ~ & ~ & {\color{blue} ($\times$1.11)} & {\color{blue} ($\times$1.03)} & {\color{blue} ($\times$1.20)} & {\color{blue} ($\times$1.03)} & {\color{blue} ($\times$1.02)} & {\color{blue} ($\times$1.04)} & {\color{blue} ($\times$1.18)} & {\color{blue} ($\times$1.15)} & {\color{blue} ($\times$1.18)} \\
        ~ & \multirow{2}{*}{8} & 31.75 & 66.29 & 71.24 & 90.61 & 224.79 & 225.71 & 172.31 & 416.19 & 529.39 \\
        ~ & ~ & {\color{blue} ($\times$1.29)} & {\color{blue} ($\times$1.15)} & {\color{blue} ($\times$1.25)} & {\color{blue} ($\times$1.12)} & {\color{blue} ($\times$1.05)} & {\color{blue} ($\times$1.10)} & ($\times$1.49) & ($\times$1.43) & ($\times$1.78) \\ \hline
        \multicolumn{2}{c|}{FP16} & 38.59 & 102.34 & 103.04 & 142.05 & 369.12 & 382.21 & 206.88 & 538.21 & 553.73 \\ \hline
    \end{tabular}
    }
\end{table*}

\end{document}